\documentclass[lettersize,journal]{IEEEtran}
\usepackage{amsmath,amsfonts}
\usepackage{amssymb}
\usepackage{algorithmic}
\usepackage{algorithm}
\usepackage{amsthm}
\usepackage{array}
\usepackage[caption=false,font=normalsize,labelfont=sf,textfont=sf]{subfig}
\usepackage{array,multirow,graphicx}
\usepackage{color,soul}
\usepackage{multirow}
\definecolor{mygray}{gray}{.9}
\usepackage{textcomp}
\usepackage{stfloats}
\usepackage{url}
\usepackage{verbatim}
\usepackage{graphicx}
\usepackage{cite}
\usepackage{graphicx}
\usepackage{amsmath}
\usepackage{amssymb}
\usepackage{booktabs}
\usepackage{mathtools}
\usepackage{bm}
\showoutput
\usepackage[table]{xcolor}
\newtheorem{corollary}{Corollary}

\newtheorem{theorem}{Theorem}

\newcommand{\be}{ \mbox{$ \bf E $}}

\newcommand{\defeq}{\mbox {$  \ \stackrel{\Delta}{=} $}}

\def\argmax{\operatornamewithlimits{arg\,max}}

\usepackage{algorithmic}        
\usepackage{algorithm}  
\usepackage{stfloats}

\begin{document}
\title{Conditional Mutual Information Constrained Deep Learning for Classification}

\author{En-Hui~Yang,~\IEEEmembership{Fellow,~IEEE,} Shayan~Mohajer~Hamidi$^*$,~\IEEEmembership{Member,~IEEE,} Linfeng Ye$^*$,~\IEEEmembership{Member,~IEEE,}  Renhao Tan, and Beverly Yang
\thanks{ $^*$ Authors contributed equally.
\\
En-Hui Yang, Shayan Mohajer Hamidi, Linfeng Ye and Renhao Tan are with the Department of Electrical and Computer
Engineering, University of Waterloo, Waterloo, ON N2L 3G1, Canada  (e-mail:
 ehyang@uwaterloo.ca; smohajer@uwaterloo.ca;  l44ye@uwaterloo.ca; cameron.tan@uwaterloo.ca). \\
Beverly Yang is with the NBK Institute of Mining Engineering, University of British Columbia, Vancouver, BC V6T 1Z4, Canada (e-mail: beverly.yang@ubc.ca).
 }}


 
\maketitle
% \IEEEtitleabstractindextext{%
\begin{abstract}
The concepts of conditional mutual information (CMI) and normalized conditional mutual information (NCMI) are introduced to measure the concentration and separation performance of a classification deep neural network (DNN) in the output probability distribution space of the DNN, where  CMI and the ratio between  CMI and NCMI represent the intra-class concentration and inter-class separation of the DNN, respectively. By using NCMI to evaluate popular  DNNs pretrained over ImageNet in the literature, it is shown that their validation accuracies over ImageNet validation data set are more or less inversely proportional to their NCMI values. Based on this observation, the standard deep learning (DL) framework is further modified to minimize the standard cross entropy function subject to an NCMI constraint, yielding CMI constrained deep learning (CMIC-DL).  A novel alternating learning algorithm is proposed to solve such a constrained optimization problem.  Extensive experiment results show that DNNs trained within CMIC-DL  outperform the state-of-the-art models trained within the standard DL and other loss functions  in the literature in terms of both accuracy and robustness against adversarial attacks. In addition, visualizing the evolution of learning process through the lens of CMI and NCMI is also advocated. 
\end{abstract}


% \IEEEdisplaynontitleabstractindextext
% % \IEEEdisplaynontitleabstractindextext has no effect when using
% % compsoc or transmag under a non-conference mode.

% \fi
% \IEEEpeerreviewmaketitle

\section{Introduction}\label{sec:intro}

\IEEEPARstart{I}{n} recent years, deep neural networks (DNNs)  have been applied in a wide range of applications, revolutionizing
fields like computer vision, natural language processing, and speech recognition \cite{lecun2015deep, goodfellow2016deep}. Typically, a DNN consists of cascaded non-linear layers that progressively produce multi-layers of representations with increasing levels of abstraction, starting from raw input data and ending with a
predicted output label. The success of DNNs is largely attributable to their ability to learn these multi-layers of representations as features from the raw data through a deep learning (DL) process.

Putting its neural architecture aside, a classification DNN is, mathematically, a mapping from raw data $x \in \mathbb{R}^{d}$ to a probability distribution $P_x$ over the set of class labels, predicting an output label $\hat{y}$ with probability $P_x (\hat{y})$. Given a pair of random variables $(X, Y)$, the distribution of which governs either a training set or testing set, where $X \in \mathbb{R}^{d}$ represents the raw data and $Y$ is the ground truth label of $X$, the prediction performance of the DNN is often measured by its error rate
   \[ \epsilon = \Pr \{ \hat{Y} \not = Y \}, \]
where $\hat{Y}$ is the label predicted by the DNN with probability $P_{X} (\hat{Y}) $ in response to the input $X$. The accuracy of the DNN is equal to $1- \epsilon$. The error rate is further upper bounded by the average of the cross entropy between the conditional distribution of $Y$ given $X$ and $P_{X}$ (see Section~\ref{sec:perf}).  To have better prediction performance, a DL process is then applied to minimize the error rate $\epsilon$ or its cross entropy upper bound \cite{lecun2015deep, goodfellow2016deep}.

Although the error rate of a DNN is its most important performance  as far as its prediction is concerned, focusing entirely on the error rate is not enough, and can actually lead to several problems. First, the error rate of a DNN depends not only on the DNN itself, but also on the governing joint distribution of $(X, Y)$. When a DNN has a small error rate for one governing joint distribution of $(X, Y)$, it does not necessarily imply that it would have a small error rate for another governing joint distribution of $(X, Y)$, especially when two distributions are quite different. This is essentially related to the well-known overfitting and robustness problems \cite{goodfellow2016deep, goodfellow2014explaining, carlini2017towards, madry2018towards}. Second, even when a DNN works well across different governing distributions of $(X, Y)$, it remains a block box to us, especially when its architecture is huge. We don't know why it works and how it works. Its error rate does not reveal any useful information about the intrinsic mapping structure such as the intra-class concentration and inter-class separation of the DNN in its output probability distribution space.

To gain deep insights into the intrinsic mapping structure of a DNN as a mapping from $x \in \mathbb{R}^{d}$ to  $P_x$, in this paper we introduce information quantities from information theory \cite{cover1999elements} to measure intra-class concentration and inter-class separation of the DNN. Specifically, we propose to use the conditional mutual information (CMI) $I(X; \hat{Y} |Y)$ between $X$ and $\hat{Y}$ given $Y$ as the measure for the intra-class concentration of the DNN as a mapping $x \in \mathbb{R}^{d} \to P_x$. For each class label $y$, the conditional mutual information $I(X; \hat{Y} |Y=y)$ between $X$ and $\hat{Y}$ given $Y=y$ tells how all output probability distributions $P_X$ given $Y=y$ are concentrated around its ``centroid'', the conditional probability distribution $P_{\hat{Y} |Y =y}$. The smaller $I(X; \hat{Y} |Y=y)$ is, the more concentrated all output probability distributions $P_X$ given $Y=y$ are  around its centroid.  We further introduce another information quantity (see Section~\ref{sec:perf}) to measure the inter-class separation of the DNN as a mapping $x \in \mathbb{R}^{d} \to P_x$. Define the ratio between  $I(X; \hat{Y} |Y)$ and the inter-class separation as the normalized conditional mutual information (NCMI)  between $X$ and $\hat{Y}$ given $Y$. One may interpret CMI and NCMI as certain mapping structure traits of the DNN. Then in addition to its error rate, the DNN can also be evaluated in terms of its CMI and NCMI.

Equipped with our new concepts of CMI and NCMI,  we further evaluate popular DNNs pretrained in the literature over ImageNet in terms of their respective CMI and NCMI. It turns out that their validation accuracies over the ImageNet validation data set are more or less inversely proportional to their NCMI values. In other words, even though these DNNs have different architectures and different sizes, their error rates and NCMI values have more or less a positive linear relationship. Indeed, the correlation between the error rate and NCMI is above $0.99 $. This implies that given a DNN architecture, one may be able to further improve the effectiveness of DL by simultaneously minimizing the error rate (or cross entropy upper bound) and NCMI of the DNN during the learning process, where the error rate and NCMI represent the prediction performance and the concentration/separation mapping structure performance of the DNN, respectively.  This in turn motivates us to modify the standard DL framework to minimize the standard cross entropy function subject to an NCMI constraint, yielding CMI constrained deep learning (CMIC-DL).  To this end, a novel alternating learning algorithm is further proposed to solve such a constrained optimization problem.  Extensive experiment results show that DNNs trained within CMIC-DL  outperform the state-of-the-art models trained within the standard DL and other loss functions  in the literature in terms of both accuracy and robustness against adversarial attacks.

The remainder of this paper is organized as follows. In Section~\ref{sec:perf}, we formally introduce the concepts of CMI and NCMI to measure intra-class concentration and inter-class separation structure performance of a DNN when it is viewed as a mapping from $x \in \mathbb{R}^{d}$ to  $P_x$. In Section~\ref{sec:eval}, we use NCMI to evaluate and compare popular DNNs pretrained in the literature over ImageNet. These DNNs have different architectures and different sizes. Section~\ref{sec:cmic-dl} is devoted to the full development of CMIC-DL. In Section~\ref{sec:exp}, extensive experiment results are presented and compared with the prior art in the literature; visualizing the evolution of learning process through the lens of CMI and NCMI is also advocated. Finally, conclusions are drawn along with some open problems in Section~\ref{sec:con}.

\section{Performance of DNNs: Concentration and Separation } \label{sec:perf}

A DNN can be described either by its neural architecture along with its connection weights, the number of which can be in billions, or by its mathematical mapping from $x \in \mathbb{R}^{d}$ to  $P_x$. Both perspectives are useful. In this and next sections, we will take the second perspective and regard a DNN simply as a mapping $x \in \mathbb{R}^{d} \to  P_x$. Before formally introducing CMI and NCMI, we set up notation to be used throughout the paper.

\subsection{Notation}
For a positive integer $K$,  let $[K]\triangleq \{1,\dots,K\}$. Assume that there are $C$ class labels with $[C]$ as the set of class labels. Let ${\cal P} ([C])$ denote the set of all probability distributions over $[C]$. For any two probability distributions $P_1, P_2 \in {\cal P} ([C])$, the cross entropy of $P_1$ and $P_2$ is defined as
  \begin{equation} \label{eq2-1}
  H (P_1, P_2 ) = \sum_{i=1}^C -P_1 (i) \ln P_2 (i) ,
  \end{equation}
where $\ln$ denotes the logarithm with base $e$; the Kullback–Leibler (KL) divergence (or relative entropy) between $P_1$ and $P_2$ is defined as
   \begin{equation} \label{eq2-2}
  D (P_1 || P_2 ) = \sum_{i=1}^C P_1 (i) \ln {P_1 (i) \over  P_2 (i)} .
  \end{equation}
For any $y \in [C]$ and $P \in {\cal P} ([C])$, write the cross entropy of the one-hot probability distribution corresponding to $y$ and $P$ as
   \begin{equation} \label{eq2-3}
  H (y, P ) =  - \ln P (y) .
  \end{equation}
Given a DNN:  $x \in \mathbb{R}^{d} \to  P_x$, let $\mathbf{\theta}$ denote its weight vector consisting of all its connection weights; whenever there is no ambiguity, we also write $P_x$ as $P_{x, \mathbf{\theta}}$,  and $P_x (y) $ as $P(y |x, \mathbf{\theta})$ for any $y \in [C]$.

%For any function $F$ of $\mathbf{\theta}$, $\nabla_\mathbf{\theta} F(\theta)$ denotes the gradient vector of function $F(\cdot)$ with respect to the vector $\mathbf{\theta}$.
%For a vector  $x= \{ x_i \} $, let $\{x_i\}_{i \in [K]}=\{x_1,\dots,x_K\}$,  the set of all components of $x$ with indices from $[K]$. Scalars are denoted by lowercase letters, e.g., $w$. We use bold-face lowercase letters and  bold-face capital letters to represent vectors (e.g., $\mathbf{w}$) and matrices (e.g., $\mathbf{W}$), respectively. Denote by $\mathbf{w}[i]$ the $i$-th element of vector $\mathbf{w}$. For two vectors $\mathbf{u}$ and $ \mathbf{v}$, denote by $\mathbf{u}\cdot \mathbf{v}$ their inner product. In addition, $\mathbb{R}$ and $\mathbb{Z}$ are the sets of real and integer values, respectively. We use $\|\mathbf{w}\|_p$ to represent the $l_p$ norm of a vector $\mathbf{w}$, and $|\mathcal{W}|$ to denote the cardinality of a set $\mathcal{W}$. Additionally, $\nabla_\mathbf{w} F(\cdot)$ denotes the gradient vector of function $F(\cdot)$ w.r.t. vector $\mathbf{w}$.

\subsection{Error Rate}
Fix a DNN:  $x \in \mathbb{R}^{d} \to  P_x$. As before, let $(X, Y)$ be a pair of random variables representing the raw input data and the corresponding ground truth label; let $\hat{Y}$ be the label predicted by the DNN with probability $P_{X} (\hat{Y})$ in response to the input $X$, that is, for any input $ x \in \mathbb{R}^{d}$ and any  $\hat{y} \in [C]$
  \begin{equation} \label{eq2-4}
  P(\hat{Y} = \hat{y} |X =x ) = P_{x} (\hat{y}) = P(\hat{y} |x, \mathbf{\theta}) .
  \end{equation}
Note that $Y \to X \to \hat{Y}$ forms a Markov chain in the indicated order. Therefore, given $X =x$, $Y$ and $\hat{Y}$ are conditionally independent.

The error rate of the DNN for $(X, Y)$ is equal to
\[ \epsilon  =  \Pr \{ \hat{Y} \not = Y \} \]
which can be upper bounded by the average of the cross entropy of the conditional probability distribution of $Y$ given $X$, $P_{Y|X} = P_{Y|X} (\cdot |X)$, and $P_X$, as shown in the following theorem. 

\begin{theorem} \label{th1}  For any DNN:  $x \in \mathbb{R}^{d} \to  P_x$ and any $(X, Y)$,
 \begin{equation} \label{eq2-4+}
\epsilon \leq \be_{X} \left [ H (P_{Y|X}, P_{X}) \right ] 
\end{equation}
where  $\be_X$ denotes the expectation with respect to $X$.    
\end{theorem}

\begin{IEEEproof} Let $I_{\{ \hat{Y} \not = Y \}}$ denote the indicator function of the event $\{ \hat{Y} \not = Y \} $. Then
\begin{eqnarray}
\epsilon & = & \Pr \{ \hat{Y} \not = Y \} \nonumber \\
          & = & \be [I_{\{ \hat{Y} \not = Y \}} ] \nonumber \\
          & = & \be_{X} \left [ \be [ I_{\{ \hat{Y} \not = Y \}} | X] \right ] \nonumber \\
          &  = & \be_{X} \left  [ 1 - \sum_{i=1}^C P_{Y|X} (i |X) P_{X} (i) \right ]  \label{eq2-5} \\
          &  = & \be_{X} \left  [  \sum_{i=1}^C P_{Y|X} (i |X) (1-  P_{X} (i) ) \right ]  \nonumber  \\
           &  \leq & \be_{X} \left [  \sum_{i=1}^C - P_{Y|X} (i |X) \ln  P_{X} (i)  \right  ]  \label{eq2-6}  \\
           & = & \be_{X} \left [ H (P_{Y|X}, P_{X}) \right ] \label{eq2-7}
\end{eqnarray}
where  \eqref{eq2-5} follows from the fact that $Y$ and $\hat{Y}$ are conditionally independent given $X$, and  \eqref{eq2-6} is due to the inequality $\ln z \leq z -1 $ for any $z >0$. This completes the proof of Theorem~\ref{th1}.
\end{IEEEproof}

Given $X=x$, what happens if the DNN outputs instead the top one label $\hat{Y}^* $
  \[ \hat{Y}^* = \argmax_{i \in [C]} P_x (i) ?\]
In this case, the  error rate of the DNN for $(X, Y)$ is equal to
\[ \epsilon^*  =  \Pr \{ \hat{Y}^* \not = Y \}\]
which can also be upper bounded in terms of $ \be_{X} \left [ H (P_{Y|X}, P_{X}) \right ]$.

\begin{corollary} \label{cor1} For any DNN:  $x \in \mathbb{R}^{d} \to  P_x$ and any $(X, Y)$,
 \begin{equation} \label{eq2-7+}
      \epsilon^* \leq C \epsilon \leq C \be_{X} \left [ H (P_{Y|X}, P_{X}) \right ]. 
 \end{equation}
\end{corollary}

\begin{IEEEproof}
 \begin{eqnarray}
 \epsilon^* & = & \Pr \{ \hat{Y}^* \not = Y \} \nonumber \\
   & = & \be_X \left [ 1 - P_{Y|X} ( \hat{Y}^* |X ) \right ] \nonumber \\
   & \leq & C \be_X \left [ P_X (\hat{Y}^*)  (1 - P_{Y|X} ( \hat{Y}^* |X )) \right ] \label{eq2-8} \\
   & \leq & C \be_{X} \left  [  \sum_{i=1}^C P_X (i) ( 1- P_{Y|X} (i |X))   \right ]  \nonumber  \\
   & = & C \be_{X} \left  [ 1 - \sum_{i=1}^C P_{Y|X} (i |X) P_{X} (i) \right ] \nonumber \\
   & = & C \epsilon \leq C \be_{X} \left [ H (P_{Y|X}, P_{X}) \right ], \label{eq2-9}
 \end{eqnarray}
 where \eqref{eq2-8} follows from the fact that $P_X  (\hat{Y}^* ) \geq 1/C$, and \eqref{eq2-9} is due to \eqref{eq2-5} and \eqref{eq2-7}.
\end{IEEEproof}

 In view of Theorem~\ref{th1} and Corollary~\ref{cor1}, no matter which form of error rate $\epsilon$ or $\epsilon^*$ is used, minimizing the average of the cross entropy $\be_X [H (P_{Y|X}, P_{X})] $ would have an effect to reduce $\epsilon$ and  $\epsilon^*$. This provides mathematical justifications for the use of the average of the cross entropy $\be_X [H (P_{Y|X}, P_{X})] $ as an objective function or a major component thereof in DL and knowledge distillation, where $ P_{Y|X}$ is approximated by the one-hot probability vector corresponding to $Y$ in DL\cite{lecun2015deep,goodfellow2016deep}, and by the output probability distribution of the teacher in knowledge distillation \cite{hinton2015distilling, nkd, menon2021statistical}. 

\begin{figure*}
\centering  \includegraphics[width=0.8\textwidth]{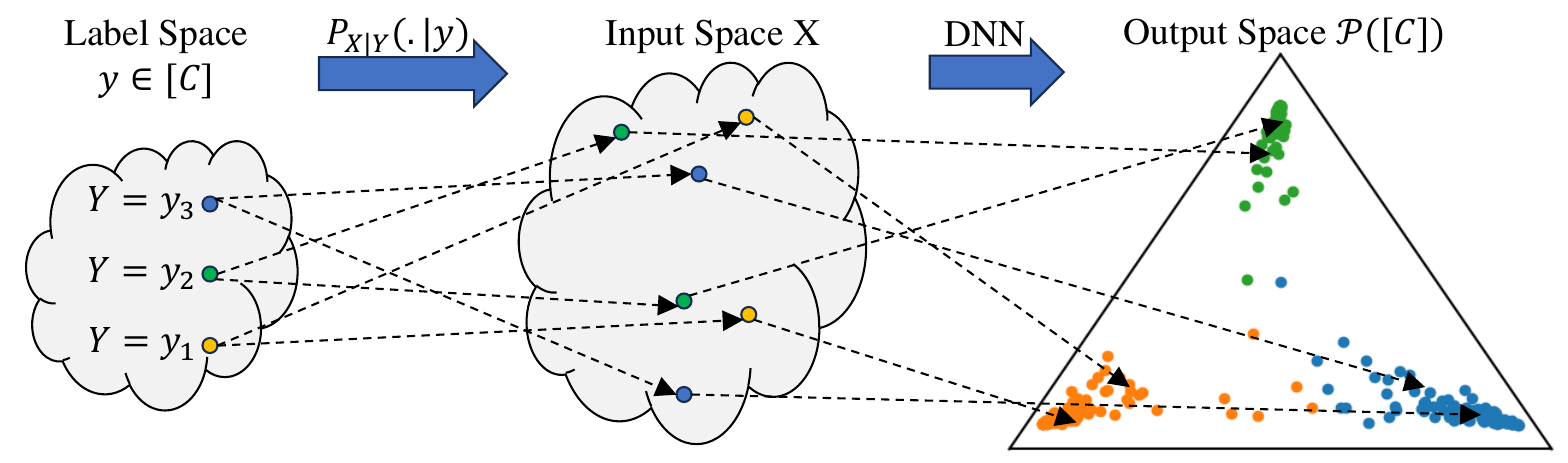}
\vspace{-2mm}
  \caption{The mappings from the label space to the input space, and from the input space to the output space of a DNN. Here caricatures are used to depict label and input spaces, where each of the three instances in the label space are mapped to two instances in input space according to $P_{X|Y}(\cdot|Y=y_i)$, for $i \in \{1,2,3\}$. On the other hand, the figure for the output space is obtained from a real example, where for the ResNet56 model trained on CIFAR-100 dataset, the output probability vectors corresponding to all validation sample instances from three randomly-picked classes 
  are projected over the two-dimensional probability simplex. 
} \label{fig:con_sep}
  % \vspace{-2mm}
\end{figure*}

\subsection{Concentration}

 The error rates $\epsilon$ and $\epsilon^*$ of the DNN:  $x \in \mathbb{R}^{d} \to  P_x$  for $(X, Y)$ do not provide any useful information on the intrinsic mapping structure  of the DNN in the probability distribution space ${\cal P}([C])$. Two important mapping structure properties the DNN: $x \in \mathbb{R}^{d} \to  P_x$ possesses, are its intra-class  concentration and inter-class separation in the space ${\cal P}([C])$. In this and next subsections, we formally introduce information quantities to quantify these two mapping structure properties, respectively.

 Visualize the  DNN: $x \in \mathbb{R}^{d} \to  P_x$ according to Fig.~\ref{fig:con_sep}. Given $Y =y$, $y \in [C]$, the input data $X$ is  conditionally distributed according to the conditional distribution $P_{X|Y} (\cdot | y)$ and then mapped into $P_X$, a random point in the space ${\cal P}([C])$. The instances (or realizations) of this random point $P_X$ form a cluster in the space ${\cal P}([C])$. The centroid of this cluster is the average of $P_X$ with respect to the conditional distribution $P_{X|Y} (\cdot | y)$, which is exactly the conditional distribution of $\hat{Y}$ given $Y=y$
   \begin{equation} \label{eq2-10}
   P_{\hat{Y} |y } = \be [ P_X | Y =y ] .
   \end{equation}
Measure the ``distance'' between each $P_X$ and the centroid $ P_{\hat{Y} | y }$ by their KL divergence $D (P_X || P_{\hat{Y} |y })$. Then the average of KL divergence $D (P_X || P_{\hat{Y} | y })$ with respect to the conditional distribution $P_{X|Y} (\cdot | y)$  is equal to
  \begin{eqnarray}
 \lefteqn{ \be \left [ D (P_X || P_{\hat{Y} |y }) | Y =y \right ] } \nonumber \\
    & = & \be \left [\left ( \sum_{i=1}^C P_X (i) \ln { P_X (i) \over P_{\hat{Y} | y } (\hat{Y} = i |Y =y )}\right ) \left  | Y=y \right.  \right ] \nonumber \\
    & = & \sum_{x} P_{X|Y} (x | y)  \left [ \sum_{i=1}^C P(\hat{Y} = i |x) \times \right. \nonumber \\
     & & \left. \ln { P (\hat{Y} =i |x ) \over P_{\hat{Y} | y } (\hat{Y} = i |Y =y )}  \right ] \label{eq2-11} \\
    & = & I (X; \hat{Y} | y ), \label{eq2-12}
  \end{eqnarray}
  where $ I (X; \hat{Y} | y ) $ is the conditional mutual information between $X$ and $\hat{Y}$ given $Y=y$. (Please refer to \cite{cover1999elements} for the notions of mutual information and conditional mutual information.) In \eqref{eq2-11}, $X$ is assumed to be discrete; if $X$ is continuous, then the average $\sum_{x} P_{X|Y} (x | y)$ should be replaced by the integral
   \[ \int_x d P_{X|Y} (x | y).  \]
  Note that \eqref{eq2-12} is due to the fact that $Y \to X \to \hat{Y}$ forms a Markov chain.

  The information quantity  $ I (X; \hat{Y} | y ) $ quantifies the concentration of the cluster formed by the instances of the random point $P_X$ given $Y =y$  around its centroid $ P_{\hat{Y} | y }$. Averaging $ I (X; \hat{Y} | y ) $ with respect to the distribution $P_Y (y)$ of $Y$, we get the conditional mutual information $I(X; \hat{Y} | Y)$ between $X$ and $\hat{Y}$ given $Y$:
   \begin{eqnarray} \label{eq:empCMI}
   I(X; \hat{Y} | Y) & = & \sum_{y \in [C]} P_Y (y) I (X; \hat{Y} | y ) \nonumber \\
    & = &  \be \left [ D (P_X || P_{\hat{Y} |Y})  \right ] \nonumber \\
    & = & \sum_{y } \sum_{x} P(x , y)  \left [ \sum_{i=1}^C P(\hat{Y} = i |x) \times \right. \nonumber \\
     & & \left. \ln { P (\hat{Y} =i |x ) \over P_{\hat{Y} | y } (\hat{Y} = i |Y =y )}  \right ]. \label{eq2-13}
   \end{eqnarray}
   The CMI $I(X; \hat{Y} | Y)$ can then be regarded as a measure for the intra-class concentration of the DNN:  $x \in \mathbb{R}^{d} \to  P_x$  for $(X, Y)$.

   In practice, the joint distribution $P(x , y)$ of $(X, Y)$ may be unknown. To compute the CMI $I(X; \hat{Y} | Y)$ in this case, one may approximate $P(x , y)$ by the empirical distribution of a data sample $\{ (x_1, y_1), (x_2, y_2), \cdots, (x_n, y_n) \}$. For any $y \in [C]$, let
    \begin{equation}
    n_y = |\{ (x_j, y_j ) : y_j = y, 1\leq j \leq n \}|, \label{eq2-14}
    \end{equation}
  where $|S|$ denotes the cardinality of a set $S$,  and
   \begin{equation}
   P_y = {1 \over n_y } \sum_{ (x_j, y_j): y_j =y } P_{x_j} . \label{eq2-15}
   \end{equation}
  Then $I(X; \hat{Y} | Y)$ can be computed as follows
   \begin{align}
  I(X; \hat{Y} | Y) 
 & =  \sum_{y \in [C]} \sum_{ (x_j, y_j): y_j =y } {1 \over n } D( P_{x_j} || P_y) \nonumber \\
 & =  {1\over n} \sum_{j=1}^n D( P_{x_j} || P_{y_j}) . \label{eq2-16}
 \end{align}

  \subsection{Separation and NCMI}

  Let $(U, V)$ be a pair of random variables independent of $(X, Y)$, and having the same joint distribution as that of $(X, Y)$. With reference to Fig.~\ref{fig:con_sep}, we define the following information quantity\footnote{Other information quantities can also be defined and used as a measure for the inter-class separation of the DNN: $x \in \mathbb{R}^{d} \to  P_x$, which will be explored in Appendix~\ref{app-sep}. Although they are more or less equivalent, the information quantity $\Gamma$ defined here is more convenient for the selection of hyper parameters in our proposed CMIC deep learning.}
   \begin{equation} \label{eq2-17}
   \Gamma = \be \left [ I_{\{Y \not = V\}} H(P_X, P_U) \right ],
   \end{equation}
  and use $\Gamma$ as a measure for the inter-class separation of the DNN: $x \in \mathbb{R}^{d} \to  P_x$. It is clear that the larger $\Gamma$ is, the further apart different clusters are from each other on average.

  Ideally, we want $I(X; \hat{Y} | Y)$ to be small while keeping $\Gamma$ large. This leads us to consider the ratio between $I(X; \hat{Y} | Y)$  and $\Gamma$:
    \begin{equation} \label{eq2-18}
  \hat{I} (X; \hat{Y} |Y ) \defeq { I(X; \hat{Y} | Y) \over \Gamma }.
  \end{equation}
We call $ \hat{I} (X; \hat{Y} |Y ) $ the normalized conditional mutual information between $X$ and $\hat{Y}$ given $Y$.

In case where the joint distribution $p(x, y) $ of $(X, Y)$ is unknown, it can be approximated by the empirical distribution of a data sample $\{ (x_1, y_1), (x_2, y_2), \cdots, (x_n, y_n) \}$. In parallel with \eqref{eq2-16}, $\Gamma$  can be computed in this case as follows:
 \begin{equation} \label{eq2-19}
  \Gamma = {1\over n^2} \sum_{j=1}^n \sum_{k=1}^n I_{\{ y_j \not = y_k \}} H(P_{x_j}, P_{x_k}),
  \end{equation}
  from which and  \eqref{eq2-16}, $  \hat{I} (X; \hat{Y} |Y )  $ can be computed accordingly.

  \subsection{Related Works}

In the literature, intra-class concentration and inter-class separation of a DNN have been mainly investigated in the feature space corresponding to the penultimate layer of the DNN, and largely treated in an ad-hoc manner in a deep learning process or algorithm. Specifically, it was observed numerically in \cite{oyallon2017building,papyan2020traces,papyan2020prevalence} that DNNs concentrate features of each class around their separated mean. This observation was further analyzed in \cite{zarka2020separation} under the Gaussian mixture model assumption about features. In \cite{wen2019comprehensive,shi2021constrained,wen2016discriminative,qi2017contrastive,ranasinghe2021orthogonal} and references therein, different loss functions including the so-called center loss, contrastive center loss, orthogonal project loss, constrained center loss, and their variants, all of which are defined in the feature space, were proposed and used in the respective learning processes to improve the intra-class concentration and inter-class separation of such trained DNNs.

In contrast, in this paper we investigate  the intra-class concentration and inter-class separation of a DNN in its output probability distribution space ${\cal P} ([C])$, where the DNN is viewed as a mapping from $x \in \mathbb{R}^{d}$  to $  P_x$. This perspective allows us to introduce information quantities, CMI, $\Gamma$, and NCMI, to quantify the intra-class concentration and inter-class separation of each DNN. In addition, our introduced CMI and NCMI can also be regarded as additional performance metrics for any DNN, which are in parallel with the error rate performance metric, are independent of any learning process, and represent mapping structure properties of a DNN. As additional performance metrics, they can be used to evaluate and compare different DNNs regardless of the architectures and sizes of  DNNs. 

Another related work in the sense of introducing information theoretic ideas into DL is the so-called coded deep learning (CDL) \cite{CDL}, where information theoretic coding ideas are embedded into the inner workings of DL. The purposes of CDL are to eliminate essentially floating operations of a coded DNN during its inference time and efficiently compress the coded DNN while maintaining or even improving the error rate of the coded DNN.

In the next section, CMI and  NCMI $  \hat{I} (X; \hat{Y} |Y )  $ will be used to evaluate and compare popular DNNs pre-trained over ImageNet in the literature.

\section{NCMI Vs. Accuracy } \label{sec:eval}

The popular DNNs we selected for evaluation according to their respective CMI and NCMI are ResNet-$\{18,34,50,101,152\}$ \cite{he2016deep}, VGG-$\{11,13,16,19\}$  \cite{simonyan2014very}, EfficientNet-$\{\text{B0},\text{B1},\text{B2},\text{B3}\}$   \cite{tan2019efficientnet}, Wide-ResNet-$\{50,101\}$  \cite{zagoruyko2016wide}, MobileNet-V3-$\{\text{small},\text{large}\}$ \cite{zhao2022new}, and AlexNet \cite{krizhevsky2012imagenet}. They are all pre-trained on ImageNet dataset and obtained from the Pytorch official website\footnote{https://pytorch.org/vision/stable/models.html.}.

 Table~\ref{tab:cmi} lists the values of CMI, $\Gamma$, and NCMI of the selected DNNs, which are calculated, according to 
 \eqref{eq2-16}, \eqref{eq2-19}, and \eqref{eq2-18}, over the ImageNet validation set, along with their respective error rate $\epsilon^*$.  From Table~\ref{tab:cmi}, it is clear that within the same family, as the model size increases, the CMI value decreases. This shows that larger models have more compact clusters in the output probability space ${\cal P} ([C])$. For the $\Gamma$ value, although the general trend is that  within the same family, the $\Gamma$ value increases as the model size gets larger, there does exist an exception. Note that for the EfficientNet family, the smallest model EfficientNet-B0 has the largest $\Gamma$ value.

\begin{figure}[!t] 
\centering\includegraphics[width=1\linewidth]{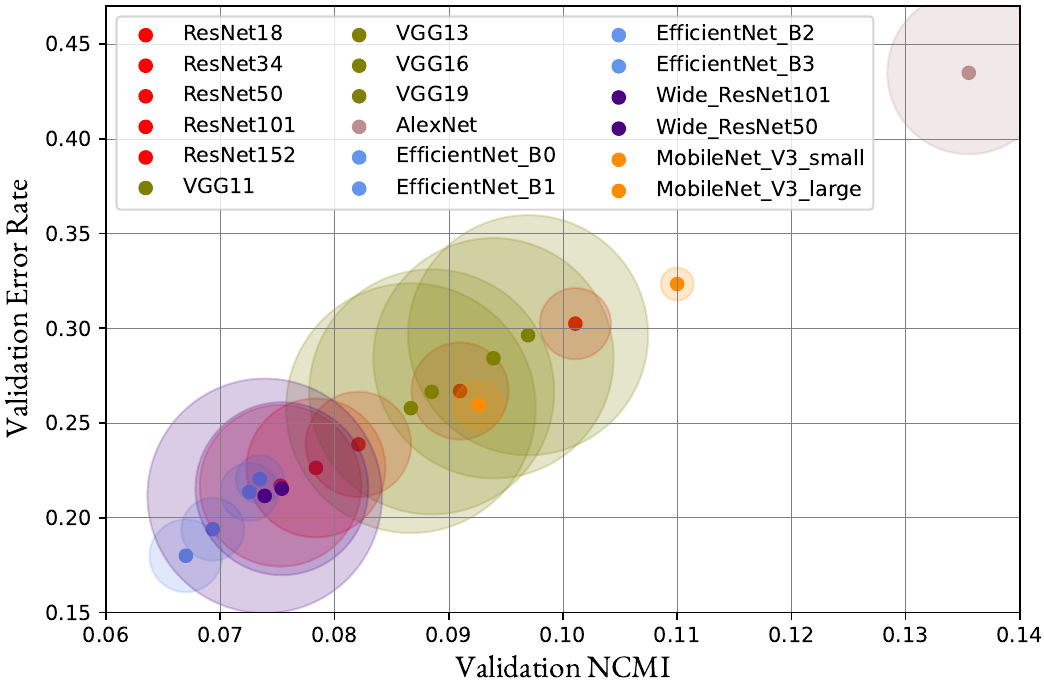}
    \caption{The error rate vs NCMI value over the validation set of popular pre-trained models on ImageNet dataset. The sizes of the circles represent the sizes of  respective models  in terms of the number of model parameters; the larger the circle, the larger the model.} 
\vskip -0.1in
    \label{fig:NCMI}
\end{figure}

 Now turn our attention to the NCMI value. From Table~\ref{tab:cmi},  it follows that as the model size within the same family increases, the NCMI value decreases as well. Even more interesting is the relationship between the NCMI and error rate $\epsilon^*$. Across all models evaluated, as the NCMI value decreases, so does the error rate $\epsilon^*$. To make the  relationship between the NCMI and error rate $\epsilon^*$ more transparent, Figure~\ref{fig:NCMI} illustrates the relationship graphically. From Figure~\ref{fig:NCMI}, it seems that the NCMI and error rate $\epsilon^*$ have a positive linear relationship; indeed,  the Pearson correlation coefficient $\rho$ \cite{cohen2009pearson} between them is  $\rho=0.9929$, strongly supporting the former statement. As such, the NCMI value of a DNN can be used to gauge the prediction performance of the DNN.

To conclude this section, let us draw some analogies. If a DNN is analogized with a student, then the error rate and NCMI of the DNN can be analogized with the testing score of the student in an exam and certain trait of the student, respectively. In a way similar to using the trait of the student to predict the student's testing performance, one can also use the NCMI value of the DNN to predict the DNN's testing performance.

\begin{table*} \label{tab:cmi}
  \centering
    \caption{CMI, $\Gamma$, and NCMI values over the validation set of some pre-trained models on ImageNet dataset along with their error rate $\epsilon^*$, where the DNNs from the same family are highlighted by the same color.}
  \resizebox{0.9\textwidth}{!}{
  \begin{tabular}{|c|c|c|c|c||c|c|c|c|c|c|}
    \hline
    Models & CMI & $\Gamma$ & NCMI & Error rate $\epsilon^*$ & Models & CMI & $\Gamma$ & NCMI & Error rate $\epsilon^*$
    % \midrule
     \\ \hline \hline
    \cellcolor{red!30}ResNet18 & \cellcolor{red!30} 0.999 & \cellcolor{red!30} 9.891 & \cellcolor{red!30} 0.101 & \cellcolor{red!30} 0.302&  \cellcolor{pink!30} AlexNet & \cellcolor{pink!30} 1.331 & \cellcolor{pink!30} 9.830 & \cellcolor{pink!30} 0.135 & \cellcolor{pink!30}  0.434  
    \\ \hline
    \cellcolor{red!30}ResNet34 & \cellcolor{red!30} 0.902 & \cellcolor{red!30} 9.919 & \cellcolor{red!30} 0.090 & \cellcolor{red!30}  0.266& 
    \cellcolor{blue!30} EfficientNet-B0 & \cellcolor{blue!30} 0.692 & \cellcolor{blue!30} 9.433 & \cellcolor{blue!30} 0.073 & \cellcolor{blue!30} 0.220
    \\ \hline 
        \cellcolor{red!30}ResNet50 & \cellcolor{red!30} 0.815  & \cellcolor{red!30} 9.929  & \cellcolor{red!30} 0.082 & \cellcolor{red!30} 0.238 &  \cellcolor{blue!30} EfficientNet-B1 & \cellcolor{blue!30}  0.661 & \cellcolor{blue!30}  9.114 & \cellcolor{blue!30}  0.072 & \cellcolor{blue!30} 0.213
    \\ \hline
        \cellcolor{red!30}ResNet101 & \cellcolor{red!30} 0.779 & \cellcolor{red!30} 9.948 & \cellcolor{red!30} 0.078  & \cellcolor{red!30}  0.226 &\cellcolor{blue!30} EfficientNet-B2 & \cellcolor{blue!30}  0.639  & \cellcolor{blue!30}  9.224 & \cellcolor{blue!30}  0.069 & \cellcolor{blue!30} 0.193
    \\ \hline
        \cellcolor{red!30}ResNet152 & \cellcolor{red!30} 0.749 &  \cellcolor{red!30} 9.953 & \cellcolor{red!30} 0.075 &  \cellcolor{red!30} 0.216 & \cellcolor{blue!30} EfficientNet-B3  & \cellcolor{blue!30} 0.627 & \cellcolor{blue!30} 9.365 & \cellcolor{blue!30} 0.067 & \cellcolor{blue!30}  0.180
    \\ \hline 
        \cellcolor{green!30}VGG11 & \cellcolor{green!30} 0.959 &  \cellcolor{green!30} 9.899 & \cellcolor{green!30} 0.096 & \cellcolor{green!30} 0.296 &\cellcolor{purple!30} Wide-ResNet50 & \cellcolor{purple!30} 0.749 & \cellcolor{purple!30} 9.935 & \cellcolor{purple!30} 0.075  & \cellcolor{purple!30}  0.215 
    \\ \hline
        \cellcolor{green!30}VGG13 & \cellcolor{green!30} 0.930 &  \cellcolor{green!30} 9.909 & \cellcolor{green!30} 0.094  & \cellcolor{green!30} 0.284 &\cellcolor{purple!30} Wide-ResNet101 & \cellcolor{purple!30} 0.734 & \cellcolor{purple!30} 9.937 & \cellcolor{purple!30} 0.073  & \cellcolor{purple!30} 0.211
    \\ \hline
        \cellcolor{green!30}VGG16& \cellcolor{green!30} 0.878 &  \cellcolor{green!30} 9.925 & \cellcolor{green!30} 0.088 & \cellcolor{green!30} 0.266 & \cellcolor{orange!30} MobileNet-V3-Small & \cellcolor{orange!30} 1.088 & \cellcolor{orange!30} 9.898 & \cellcolor{orange!30} 0.110  & \cellcolor{orange!30} 0.323
    \\ \hline    
    \cellcolor{green!30}VGG19 & \cellcolor{green!30} 0.860 & \cellcolor{green!30}  9.930 & \cellcolor{green!30} 0.086 &  \cellcolor{green!30} 0.257 &\cellcolor{orange!30} MobileNet-V3-Large & \cellcolor{orange!30}  0.922 & \cellcolor{orange!30}  9.956 & \cellcolor{orange!30}  0.092 & \cellcolor{orange!30} 0.259
    \\ \hline   
  \end{tabular}
  }\vspace{1mm}
  \label{tab:cmi}
  \vspace{-5mm}
\end{table*}

\begin{figure*}[!t]
\begin{align}  
J_{\mathcal{B}}  \left(\lambda, \beta, \theta, \{Q_c\}_{c \in [C]} \right) &= \frac{1}{|\mathcal{B}|} \sum_{(x, y) \in \mathcal{B}} H (y, P_{x, \mathbf{\theta}})+\lambda {1\over|\mathcal{B}|} \sum_{(x, y) \in \mathcal{B}} D( P_{x,  \mathbf{\theta}} || Q_y) - \beta \frac{1}{|\mathcal{B}|^2} \sum_{ (x, y), (u, v) \in \mathcal{B}} I_{\{ y  \not = v \}} H(P_{x,  \mathbf{\theta}}, P_{u,  \mathbf{\theta}}) .\label{eq:joint_opt2}
\end{align}
\vskip -0.1in
\end{figure*}

% \begin{figure*}[b]
% \begin{equation}  \label{eq:joint_opt}
% \min_{\theta} \frac{1}{|\mathcal{B}|} \sum_{x_j \in \mathcal{B}} H (P_{Y|x_j}, P_{x_j})+\lambda {1\over|\mathcal{B}|} \sum_{x_j \in \mathcal{B}} D( P_{x_j} || P_{y_j}) - \lambda^\prime \frac{1}{|\mathcal{B}|^2} \sum_{x_j, x_k \in \mathcal{B}} I_{\{ y_j \not = y_k \}} H(P_{x_j}, P_{x_k})
% \end{equation}
% \end{figure*}

% \begin{figure*}[b]
% \begin{equation}  \label{eq:joint_opt2}
% J_{\mathcal{B}}  \left(\lambda, \beta, \theta, \{Q_c\}_{c \in [C]} \right) 
% \end{equation}
% \end{figure*}

\section{CMIC Deep Learning} \label{sec:cmic-dl}

The discussions in the above section suggest a new way of learning. In the learning process, instead of minimizing the average of cross entropy $ \be_{X} \left [ H (P_{Y|X}, P_{X}) \right ]$ alone, one also needs to look after the NCMI $\hat{I} (X; \hat{Y} |Y ) $. This leads to a new form of learning framework dubbed CMI constrained deep learning (CMIC-DL), which is described next.

\subsection{Optimization Problem Formulation} \label{opf}
In CMIC-DL, the optimization problem to be solved is as follows: 
\begin{align} \label{eq:constrained}
\min_{\mathbf{\theta}}~ &\be_{X} \left [ H (P_{Y|X}, P_{X,  \mathbf{\theta}}) \right ] \nonumber \\
\text{s.t.}~~ &\hat{I} (X; \hat{Y} |Y )=r,
\end{align}
where  $r$ is a positive constant. By interpreting $ \hat{I} (X; \hat{Y} |Y ) $ as a rate, and $ \be_{X} \left [ H (P_{Y|X}, P_{X, \mathbf{\theta} }) \right ]  $ as a distortion, the above optimization problem resembles the rate distortion problem in information theory \cite{cover1999elements, berger1971rate, yang1997fixed}. 
By rewriting the constraint in \eqref{eq:constrained}, and using the Lagrange multiplier method, the constrained optimization problem in \eqref{eq:constrained} could be formulated as the following unconstrained one
\begin{align} \label{eq:uncons1}
\min_{\mathbf{\theta} }~ &\be_{X} \left [ H (P_{Y|X}, P_{X,  \mathbf{\theta}}) \right ] \nonumber \\
&+\lambda I(X; \hat{Y} | Y)-\beta \be  \left [ I_{\{Y \not = V\}} H(P_{X,  \mathbf{\theta}}, P_{U, \mathbf{\theta}} ) \right ],
\end{align}
where $\lambda >0 $ is a scalar, and $\beta=\lambda r$.

Note that in view of \eqref{eq2-13}, the CMI $ I(X; \hat{Y} | Y) $ in \eqref{eq:uncons1} depends on $P_{\hat{Y} |Y}$, which, for $Y=y$, is the average of $P_{X,  \mathbf{\theta}}$ with respect to the conditional distribution $P_{X|Y} (\cdot |y)$ (see \eqref{eq2-10}). As such, the unconstrained optimization problem in its form \eqref{eq:uncons1} is not amenable to numerical solutions. To overcome this, we first convert it into a double unconstrained minimization problem by introducing a dummy distribution $Q_y \in {\cal P} ([C])$ for each $y \in [C]$, as shown in the following theorem, which will be proved in Appendix~\ref{app-th2}. 

\begin{theorem} \label{th2}
For any $\lambda >0 $ and $\beta >0$, 
\begin{align}  \label{eq:joint_opt}
\min_{\mathbf{\theta} }~ & \left \{ \be_{X} \left [ H (P_{Y|X}, P_{X,  \mathbf{\theta}}) \right ] \right.  \nonumber \\
& \left. + \lambda I(X; \hat{Y} | Y)-\beta \be  \left [ I_{\{Y \not = V\}} H(P_{X,  \mathbf{\theta}}, P_{U, \mathbf{\theta}} ) \right ] \right \} \nonumber \\
= & \min_{\mathbf{\theta} }~ \min_{\{Q_c\}_{c \in [C]}}~ \left \{ \be [ H (P_{Y|X}, P_{X, \mathbf{\theta}}) +\lambda D (P_{X, \mathbf{\theta}}  || Q_Y) ]  \right. \nonumber \\
& \left.  - \beta \be [I_{\{Y \not = V\}} H(P_{X, \mathbf{\theta}}, P_{U, \mathbf{\theta}}) ]  \right \} .
\end{align} 
\end{theorem}
% \begin{IEEEproof}
% See Appendix \ref{app-th2}.
% \end{IEEEproof}
In practice, the joint distribution $P(x, y)$ of $(X, Y)$ may be unknown. In this case, to solve \eqref{eq:joint_opt} numerically, one may approximate $P(x, y)$ by the empirical distribution of a data sample (such as a mini-batch in the DL process) $ \mathcal{B} = \{ (x_{i_1}, y_{i_1}), (x_{i_2}, y_{i_2}), \cdots, (x_{i_m}, y_{i_m}) \}$, and $P_{Y|X}$ by the one-hot probability distribution corresponding to $Y$. Accordingly, the objective function in the double minimization \eqref{eq:joint_opt} can be approximated by $J_{\mathcal{B}}  \left(\lambda, \beta, \theta, \{Q_c\}_{c \in [C]} \right)$ shown in  \eqref{eq:joint_opt2} (on the top of the page). 

\subsection{Algorithm for Solving the Optimization in \eqref{eq:joint_opt}} \label{subsec:alg}

Having addressed how to approximate the objection function in the double minimization \eqref{eq:joint_opt}, we are now ready to present an algorithm for solving 
 \eqref{eq:joint_opt}.  In fact, by reformulating the single minimization problem as a double minimization problem, Theorem~\ref{th2} lends us an alternating  algorithm that optimizes $\mathbf{\theta} $ and $\{Q_c\}_{c \in [C]}$ alternatively to minimize the objective function in \eqref{eq:joint_opt}, given that the other is fixed. 

Given  $\{Q_c\}_{c \in [C]}$, $\mathbf{\theta} $ can be updated using the same strategy as in the conventional DL through stochastic gradient descent iterations over mini-batches, where the training set is divided into $B$ mini-batches $\{\mathcal{B}_b\}_{b \in [B]}$  with each batch of size $|\mathcal{B}|$. Given $\mathbf{\theta} $, how is $\{Q_c\}_{c \in [C]}$ updated? This is where differences arise. In view of \eqref{eq2-10} and \eqref{eq4-3}, the optimal $\{Q_c\}_{c \in [C]}$ given $\mathbf{\theta} $ is equal to 
\begin{equation} \label{eq4-4}
    Q_c = P_{\hat{Y} |y=c} = \sum_{x} P(x |y=c ) P_{x, \mathbf{\theta}},
\end{equation}
for any $c \in [C]$. Therefore, to update $\{Q_c\}_{c \in [C]}$ given $\mathbf{\theta} $,  we construct, at each iteration,  $C$ mini-batches $\{\mathfrak{B}_c\}_{c \in [C]}$ in the following manner: to make $\mathfrak{B}_c$,  $\forall c \in [C]$, we randomly sample $|\mathfrak{B}_c|$ instances from the training samples whose ground truth labels are $c$. It then follows from \eqref{eq4-4} that for any $c \in [C]$, $Q_c$ is updated as\footnote{To update $\{Q_c\}_{c \in [C]}$, we may use momentum to make the updation more stable and less noisy.}
   \begin{equation} \label{eq4-5}
   Q_c = \frac{\sum_{x \in \mathfrak{B}_c} P_{x,\mathbf{\theta} }}{|\mathfrak{B}_c|} .
   \end{equation}

The procedure for solving the optimization
problem \eqref{eq:joint_opt} is now summarized in Algorithm \ref{alg:01}, where we use $(\cdot)^t_{c,b}$ to indicate class $c$ at the $b$-th batch updation during the $t$-th epoch. We also use $(\cdot)^t_{c,B}$ as $ (\cdot)^t_{c}$ whenever necessary, and set $(\cdot)^t_{c,0}=(\cdot)^{t-1}_{c}$.

\begin{algorithm}[t!]
\caption{The proposed alternating algorithm for solving the optimization problem in \eqref{eq:joint_opt}} 
\begin{algorithmic}[1] \label{alg:01}
\renewcommand{\algorithmicrequire}{\textbf{Input:}}
\renewcommand{\algorithmicensure}{\textbf{Output:}}
\REQUIRE The training set $\mathcal{T}$, all mini-batches $\{\mathcal{B}_b\}_{b \in [B]}$, number of epochs $\mathit{T}$, $\lambda$, and $\beta$.
%\ENSURE Optimal values for $\hat{\boldsymbol{\theta}}_l$, $q_l$.\\
\STATE \textbf{Initialization:}\\Initialize  $\theta^0$ and $\{Q_c^0\}_{c \in [C]}$.
\FOR {$t=1$ to $T_{max}$}
% \FOR {$\forall \mathcal{B} \in \{\mathcal{B}_b\}_{b \in [B]}$}
\FOR {$b=1$ to $B$}
\STATE \emph{$[$Updating $\theta$$]$}:

 Fix $\{Q_{c,b-1}^t\}_{c \in [C]}$. 
% Set $\{\hat{\boldsymbol{\theta}}^t_l\}_{l \in [L]}$
% Find the optimal $\{\hat{\boldsymbol{\theta}}^t_l\}_{l \in [L]}$ that achieves the following minimum

Update $\theta^t_{b-1}$ to $\theta^t_{b}$ by using (stochastic) batch gradient descent over the loss function $J_{\mathcal{B}_b}  \left(\lambda, \beta, \theta^t_{b-1}, \{Q_{c,b-1}^t\}_{c \in [C]} \right)$.

\STATE \emph{$[$Updating $\{Q_c\}_{c \in [C]}$$]$}:

Fix $\theta^t_{b}$.

Construct mini-batches $\{\mathfrak{B}_c\}_{c \in [C]}$ from $\mathcal{T}$.

Update $Q_{c,b-1}^t$ to $Q_{c,b}^t$, $\forall c \in [C]$, according to \eqref{eq4-5}, i.e., 
% \begin{align} \label{eq:optQ}
% \min_{Q_c} J_{\mathfrak{B}_c}  \left(\lambda, \beta, \theta^t_{b}, \{Q^t_{m,b-1}\}_{m \in [C]/c}, Q_c \right) . 
% \end{align}
\begin{align} \label{eq:optQ}
 Q_{c,b}^t  = \frac{\sum_{x \in \mathfrak{B}_c} P_{x,\mathbf{\theta}^t_b }}{|\mathfrak{B}_c|}.
\end{align}
\ENDFOR

\ENDFOR
\RETURN model parameters $\theta^T$.
\end{algorithmic} 
\end{algorithm}

\section{Experiment Results} \label{sec:exp}

To demonstrate the effectiveness of CMIC-DL and compare it with some state-of-the-art alternatives, we have conducted a series of experiments. 
Specifically, we have performed experiments on two popular image classification datasets, namely CIFAR-100 \cite{krizhevsky2009learning} and ImageNet \cite{krizhevsky2012imagenet}. In Subsections \ref{sec:cifar} and \ref{sec:imagenet},  we present their respective accuracy results. 
In Subsection \ref{sec:outputspace}, we explore how to visualize the concentration and separation of a DNN, which is made possible by viewing the DNN as a mapping from  $x \in \mathbb{R}^{d}$ to  $ P_x$; using such a visualization method, the concentration and separation of ResNet-56 trained within our CMIC-DL framework are then compared with those of ResNet-56 trained within the standard DL framework.

In the literature, a deep learning process is typically analyzed experimentally through the evolution curve of its error rate. With our newly introduced performance metrics, CMI, $\Gamma$ (separation), and NCMI, the learning process can also be analyzed through the evolution curves of CMI, $\Gamma$, and NCMI, which show interestingly how the mapping structure in terms of CMI, $\Gamma$, and NCMI evolves over the course of learning process. 
In Subsection \ref{sec:abl}, we use ResNet-56 as an example, and illustrate and compare the evolution curves of CMI, $\Gamma$, NCMI, and error rate within our CMIC-DL framework vs within the standard DL framework.  Lastly, in Subsection \ref{sec:robustness}, we evaluate the robustness of models trained within our CMIC-DL framework against two different adversarial attacks, and show that in comparison with the standard DL, CMIC-DL improves the robustness of DNNs as well. 

\subsection{Experiments on CIFAR-100} \label{sec:cifar}
CIFAR-100 dataset contains 50K training and 10K test colour images of size $32\times32$, which are labeled for 100 classes.

 $\bullet$ \textbf{Models}: 
To show the effectiveness of CMIC-DL, we have conducted experiments on three different model architectural families.  Specifically, we have selected (i) three models from ResNet family \cite{he2016deep}, namely ResNet-$\{32,56,110\}$; (ii) VGG-13  from VGG family \cite{simonyan2014very}; and (iii)  Wide-ResNet-28-10 from Wide-ResNet family \cite{zagoruyko2016wide}.

 $\bullet$ \textbf{Benchmarks}: We evaluate the performance of the DNNs trained via CMIC-DL against those trained by conventional cross entropy loss (CE), center loss (CL) \cite{wen2016discriminative} which promotes clustering the features, focal loss (FL) \cite{lin2017focal} which uses regularization, large-margin Gaussian Mixture (L-GM) loss \cite{wan2018rethinking} which imposes margin constraints, and orthogonal projection loss (OPL) \cite{ranasinghe2021orthogonal} which imposes orthogonality in the feature space.

 $\bullet$ \textbf{Training settings}: 
We have deployed an SGD optimizer with a momentum
of 0.9, a weight decay of 0.0005, and a batch size of 64. We have trained the models for 200 epochs, and adopted an initial learning rate of 0.1, which is further divided by
10 at the 60-th, 120-th and 160-th epochs. To have a fair comparison,  we have reproduced the results of all the benchmark methods using their respective best hyper-parameters reported in their original papers. In addition, in Algorithm \ref{alg:01}, we set $\{Q_c^0(i)\}_{c \in [C]}=\frac{1}{C}$, for $i \in [C]$, use $|\mathfrak{B}_c|=8$, $\forall c \in [C]$, and also update $Q_{c,b}^t$ using the momentum of 0.9999.

The results are summarized in Table \ref{table:cifar100}. As seen, the models trained within our CMIC-DL framework outperform those trained by the benchmark methods. Importantly, the improvement is consistent across the models from different architectural families,  showing that CMIC-DL can effectively train DNNs from different families. As a rule of thumb, compared to the CE method, CMIC-DL yields DNNs with almost 1.3\% higher validation accuracy for the ResNet models. 

Furthermore, in Table \ref{table:cifarNCMI} we report the NCMI values $\hat{I} (X; \hat{Y} |Y )$, over the validation set, for the models we trained in Table \ref{table:cifar100}, where we use the notation $\hat{I}_{\text{Loss}}$ to denote the NCMI value when the underlying DNN is trained using ``Loss'' method. As observed, $\hat{I}_{\text{CMIC}}$ has the smallest value compared to the other counterparts.  

In addition, in Table \ref{table:hype1}, we report the $\lambda^*$ and $\beta^*$ values for which we obtained the best validation accuracies. As observed, the $\lambda^*$ and $\beta^*$ values are almost the same for all the models. 

\begin{table}[!t]
\caption{ The validation accuracies (\%) of different models trained by CMIC-DL and different benchmark methods over \textbf{CIFAR-100 dataset}, which are averaged over three different random seeds, and where Bold and underlined values denote the best and second best results, respectively.} 
	\small
	\begin{center}
    \begin{tabular}{l|c|c|c|c|c}
    \toprule[0.4mm]
\rowcolor{mygray}    Loss  & Res32 & Res56 & Res110 &VGG13&  WRN-28-10 \\ \midrule
    CL
  & 70.23 & 72.70& 74.20
& 74.50
& 80.97
\\
    FL
  & \underline{71.62} &  \underline{73.20} & 74.35 & 74.53 & 81.24\\
    LGM
  & 71.50 & 73.06 &  \underline{74.39} &   \underline{74.57}  &  \underline{81.29}\\
    OPL
  & 71.03 & 72.60 & 73.98 & 74.11 & 81.12\\
   
\midrule
    CE 
  & 70.90
 & 72.40
 & 73.79 & 73.77
& 80.93
\\
CMIC
  & \textbf{72.24} &\textbf{73.66}& \textbf{75.08} & \textbf{74.62}
& \textbf{81.63}
\\  
 \bottomrule[0.4mm]
    \end{tabular}
	\end{center}
 \vspace{-1.5em}
	\label{table:cifar100}
\end{table}

\begin{table}[!t]
\caption{The respective NCMI values, measured over the validation set, of the models trained in Table \ref{table:cifar100} via different benchmark methods. The values are averaged over thee different runs.} 
	\small
	\begin{center}
    \begin{tabular}{l|c|c|c|c|c}
    \toprule[0.4mm]
\rowcolor{mygray} Loss  & Res32 & Res56 & Res110 &VGG13&  WRN-28-10 \\ \midrule
     $\hat{I}_{\text{CL}}$ & 0.057 & 0.045& 0.0395 & 0.0395 & 0.0309
\\
    $\hat{I}_{\text{FL}}$ & 0.053 & 0.046& 0.0393 & 0.0399 & 0.0312 \\
$\hat{I}_{\text{LGM}}$ & 0.054 & 0.047& 0.0390 & 0.0398 & 0.0310 \\
$\hat{I}_{\text{OPL}}$ & 0.056 & 0.050& 0.0397& 0.0402 & 0.0314 \\
\midrule
$\hat{I}_{\text{CE}}$ & 0.057 &  0.053 & 0.0402 & 0.0408 & 0.0317
\\
$\hat{I}_{\text{CMIC}}$ & 0.051 & 0.042 & 0.0382 & 0.0392 & 0.0303 \\
 \bottomrule[0.4mm]
    \end{tabular}
	\end{center}
 \vspace{-1.5em}
	\label{table:cifarNCMI}
\end{table}

\begin{table}[!t]
\centering
\caption{\textbf{Hyper-parameters}, $\lambda^*$ and $\beta^*$,  that were used in CMIC-DL in Table \ref{table:cifar100}.} 
 \resizebox{1\columnwidth}{!}{
    \begin{tabular}{l|c|c|c|c|c}
    \toprule[0.4mm]
\rowcolor{mygray}    Params.  & Res32 & Res56 & Res110 &VGG13 &  WRN-28-10 \\ \midrule
    ($\lambda^*$,$\beta^*$) & (0.7,0.4) & (0.7,0.4) & (0.7,0.2)  & (0.8,0.3) & (0.7,0.4)
\\
 \bottomrule[0.4mm]
    \end{tabular}}
 \vspace{-1.5em}
	\label{table:hype1}
\end{table}

\subsection{Experiments on ImageNet} \label{sec:imagenet}

ImageNet is a large-scale dataset used in visual
recognition tasks, containing around 1.2 million training
samples and 50,000 validation images.

$\bullet$ \textbf{Models}: 
We have conducted experiments on two models from ResNet family, namely ResNet-18 and ResNet-50.

 $\bullet$ \textbf{Benchmarks}: We evaluate the performance of CMIC-DL against CE and OPL.

 $\bullet$ \textbf{Training settings}: 
We have deployed an SGD optimizer with a momentum
of 0.9, a weight decay of 0.0001, and a batch size of 256. We have trained the models for 90 epochs, and adopted an initial learning rate of 0.1, which is further divided  by 10 at the 30-th and 60-th epochs. In Algorithm \ref{alg:01}, we set $\{Q_c^0(i)\}_{c \in [C]}=\frac{1}{C}$, for $i \in [C]$, use $|\mathfrak{B}_c|=8$, $\forall c \in [C]$, and also update $Q_{c,b}^t$ using the momentum of 0.9999.

The top-$\{1,5\}$ accuracies are reported in Table \ref{table:baseline_imagenet}. As seen, in comparison with the CE method, CMIC-DL increases the top-1 validation accuracy for ResNet-18 and ResNet-50 by 0.56\% and 0.37\%, respectively. The improvement is also consistent for the top-5 validation accuracy.

 The hyper parameters $(\lambda^*,\beta^*)$ used in CMIC-DL for ResNet-18 and ResNet-50 are $(0.6,0.1)$ and $(0.6,0.2)$, respectively. The corresponding NCMI values are  $\hat{I}_{\text{CE}} =0.110$ and $\hat{I}_{\text{CMIC}}=0.102$ for ResNet-18, and  $\hat{I}_{\text{CE}}=0.091$ and $\hat{I}_{\text{CMIC}}=0.088$ for ResNet-50.

\begin{table}
\caption{The validation accuracies (\%) of different models trained by CMIC-DL and different benchmark methods on \textbf{ImageNet dataset}.}
	\small
	\begin{center}
\begin{tabular}{l|c|c|c|c}
\toprule[0.4mm]

\rowcolor{mygray} 
\cellcolor{mygray} & \multicolumn{2}{c|}{\cellcolor{mygray}ResNet-18}                 & \multicolumn{2}{c}{\cellcolor{mygray}ResNet-50}                 \\ \cline{2-5} 
\rowcolor{mygray} 
\multirow{-2}{*}{\cellcolor{mygray}Method} & top-1   & top-5   & top-1   & top-5   \\ \midrule
CE (Baseline) & 69.91  & 89.08  & 76.15& 92.87\\
OPL & 70.27& 89.60& 76.32& 93.09\\
CMIC & \textbf{70.47}  & \textbf{89.96}  & \textbf{76.52} & \textbf{93.44} \\\bottomrule[0.4mm]
\end{tabular}
	\end{center}
	\label{table:baseline_imagenet}
\end{table}

\subsection{Concentration and Separation Visualization} \label{sec:outputspace}

In this subsection, we explore how to visualize concentration and separation of a DNN. 
Consider the data set CIFAR-100.  To visualize concentration and separation of a DNN  in a dimension reduced probability space, we randomly select three class labels. Restrict ourselves to a subset consisting of all validation sample instances with labels from the three selected labels. Given a DNN, feed each validation sample instance from the subset into the DNN, keep only three logits corresponding to the three selected labels, and then convert these three logits into a 3 dimension probability vector through the softmax operation. Following these steps in the indicated order, the DNN then maps each validation sample instance from the subset into a 3 dimension probability vector. Further project the 3 dimension probability vector into the 2 dimension simplex. Then the concentration and separation properties of the DNN for the three selected classes can be more or less visualized through the projected 2 dimension simplex. 

\begin{figure}[!t]
\centering\includegraphics[width=0.7\linewidth]{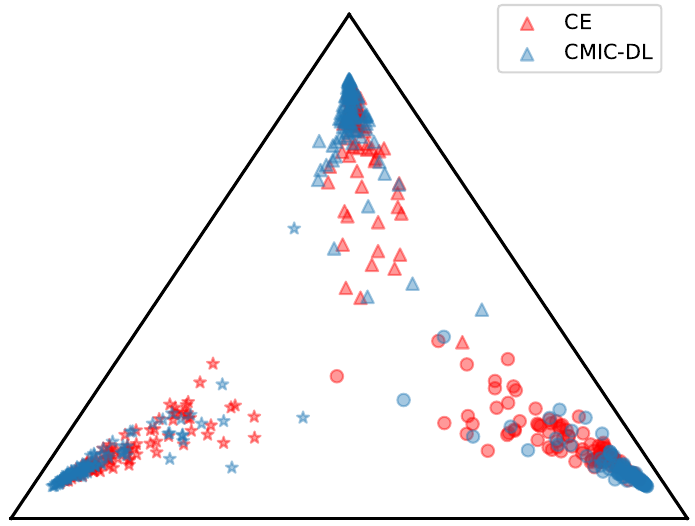}
    \caption{Visualization and comparison of concentration and separation: ResNet56 trained via CE vs  ResNet56 trained via CMIC, where different shapes indicate different classes.} 
\vskip -0.1in
    \label{fig:prob}
\end{figure}

Using the above visualization method,  Fig. \ref{fig:prob} compares the concentration and separation properties of ResNet-56 trained within our CMIC-DL framework with those of  ResNet-56 trained within the standard CE framework. From Fig. \ref{fig:prob}, it is clear that the three clusters in the case CMIC-DL are more concentrated than their counterparts in the case of CE, and also further apart from each other than their counterparts in the case of CE. Again, this is consistent with the NCMI values reported in Table~\ref{table:cifarNCMI}.

\subsection{Evolution of CMI, $\Gamma$, NCMI, and error rate} \label{sec:abl}

\begin{figure*}
  \centering 
  \subfloat[CMI ($I$).]{\includegraphics[width=0.25\linewidth]{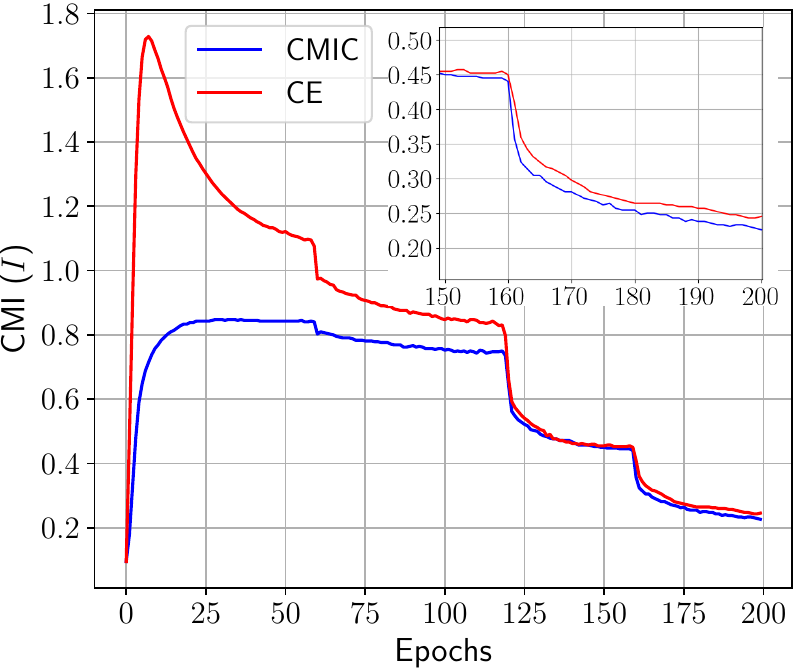}
\label{fig:cmi}}
  % \hfill
  \subfloat[Separation ($\Gamma$).]{\includegraphics[width=0.245\linewidth]{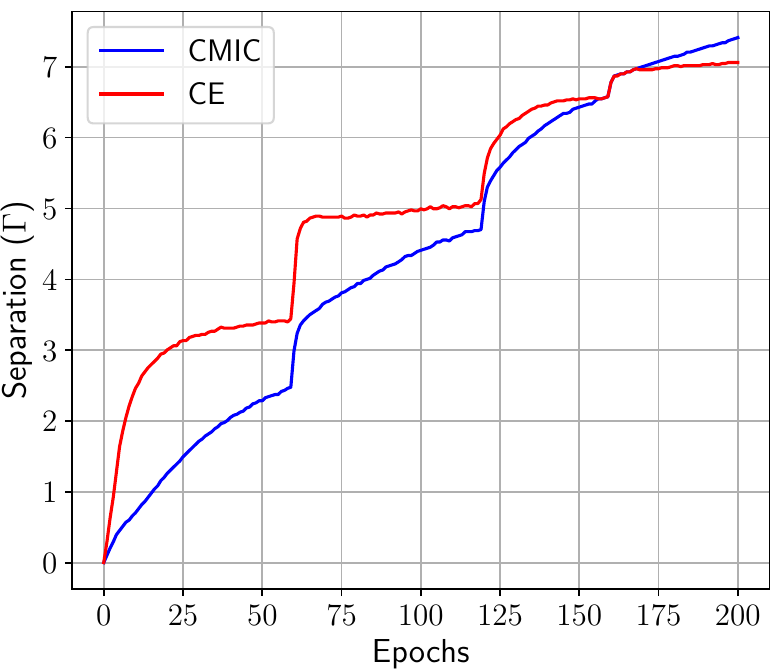}
\label{fig:sep}}
  % \hfill
  \subfloat[NCMI ($\hat{I}$).]{\includegraphics[width=0.255\linewidth]{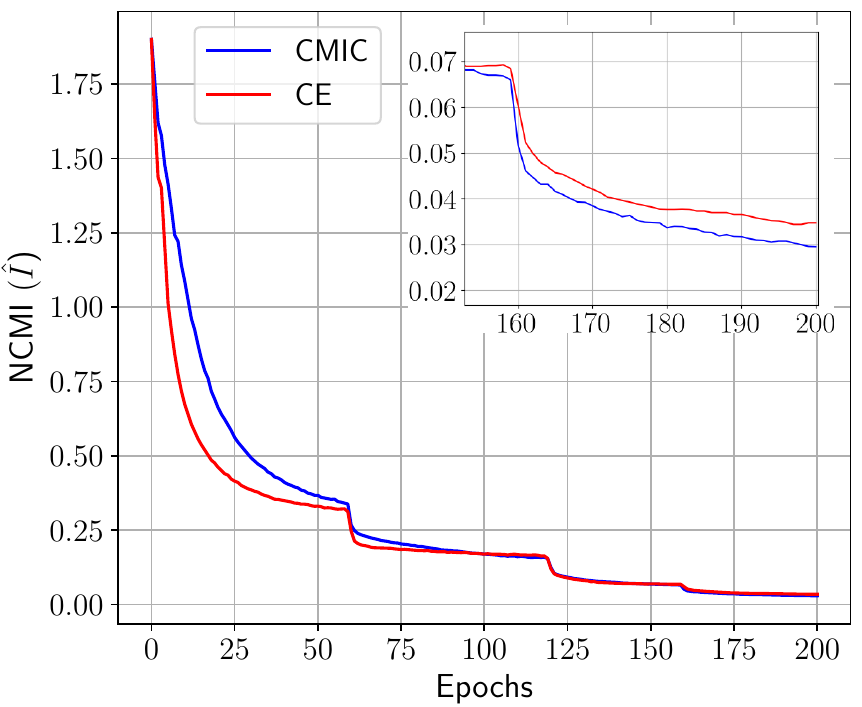}
\label{fig:ncmi}}
  \subfloat[Error rate ($\epsilon^*$).]{\includegraphics[width=0.25\linewidth]{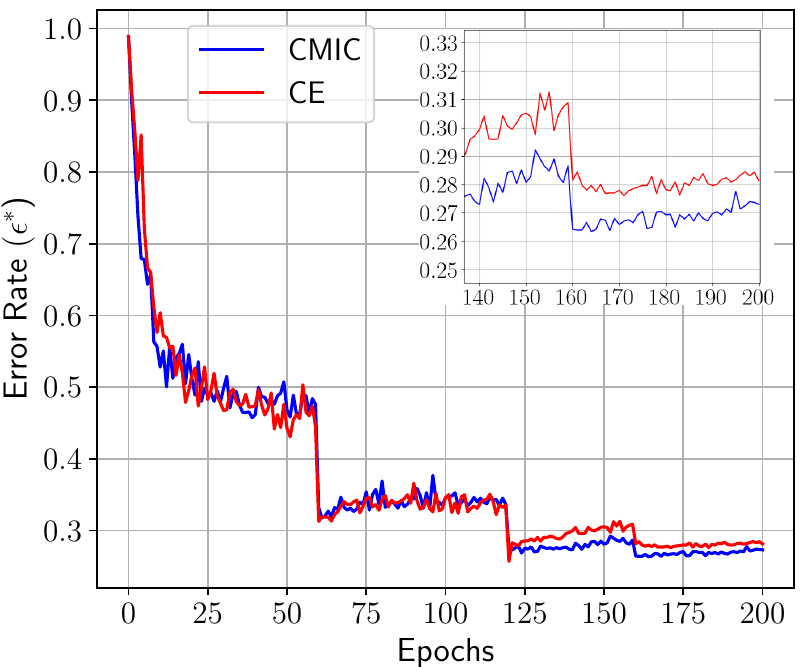}
\label{fig:error}}
  \caption{The evolution curves of (a) CMI, (b) $\Gamma$, (c) NCMI, and (d) error rate over the course of training ResNet-56  over CIFAR-100 dataset using CE and CMIC frameworks.}
  \label{fig:evolution}
  \vspace{-3mm}
\end{figure*}

In this subsection, we analyze and visualize a learning process within either our CMIC-DL framework or the conventional CE-based DL framework through the lens of   CMI, $\Gamma$, NCMI, and error rate.  Fig.~\ref{fig:evolution} shows the evolution curves of CMI, $\Gamma$,  NCMI, and error rate  over the validation set during the course of training ResNet-56 on CIFAR-100 dataset in each case, where the training setup is the same as that used in Subsection \ref{sec:cifar}, and we use $\lambda=0.7$ and $\beta=0.4$ in the case of  CMIC-DL.

As seen in Fig. \ref{fig:cmi}, the CMI value in both  CE and CMIC-DL cases is small at the beginning of the training (epoch zero). This is because at the beginning, all clusters in the output probability distribution space $\cal{P}([C])$ stick around together, as shown  from the separation distance curve (see Fig. \ref{fig:sep}), and probability distributions within each cluster  are not separated at all.  After the training starts and for the first a few epochs, the clusters move away from each other; during the course of movement, probability distributions within each cluster move in different speed, and become separated. As such, both the values of CMI and $\Gamma$ increase. Indeed, this is shown in Fig. \ref{fig:cmi} and Fig. \ref{fig:sep}.  Hereafter, the clusters continue to move away from each other, while at the same time, probability distributions within each cluster tend to move together. Thus the $\Gamma$ value continues to increase, while the CMI value decreases, as shown again in Fig. \ref{fig:cmi} and Fig. \ref{fig:sep}. 

The above summarizes the general behaviour of the CMI and  $\Gamma$ evolution curves in both CE and CMIC-DL cases. Let us now examine the differences between them. From Fig. \ref{fig:cmi}, it is clear that the CMI evolution curve in the case of CMIC-DL always  remains below its counterpart in the CE case. On the other hand, as shown in Fig. \ref{fig:sep},  although initially the $\Gamma$ value increases faster in the CE case than in the CMIC-DL case,  after the first a few epochs, the rate of increase in $\Gamma$ value is consistently higher in the CMIC-DL case than in the CE case to the extent that the $\Gamma$ value in the CMIC-DL case surpasses its counterpart in the CE case in the late stage of the learning process. 

From Fig.~\ref{fig:ncmi} and Fig.~\ref{fig:error}, we can see that once the learning process is more or less stabilized, both the NCMI value and error rate in the CMIC-DL case are consistently smaller than their counterparts in the CE case. Once again, this is consistent with our observation in Fig. \ref{fig:NCMI}: the smaller the NCMI value, the lower the error rate. In conjunction with the visualization method discussed in Subsection~\ref{sec:outputspace}, we have created a video available at https://youtu.be/G0fDwv6o9Ek to illustrate the learning process during the course of training ResNet-56 on CIFAR-100 dataset  in each of the CE and NMIC-DL cases through the lens of CMI and $\Gamma$, where concentration and separation are shown for three randomly selected classes, and the evolution curves of CMI and $\Gamma$ are shown for all classes. 

% \begin{table}[!t]
% \caption{\textbf{Effect of $\lambda$ and $\beta$ in CMIC-DL.} The validation accuracy (\%) for ResNet-56 trained on CIFAR-100 using different $\lambda$ and $\beta$ values.  The reported results are obtained via averaging over three different random seeds.} 
% 	\small
% 	\begin{center}
%     \begin{tabular}{l|c|c|c|c}
%     \toprule[0.4mm]
% \rowcolor{mygray}    Params $\lambda$-$\beta$ & 0.5-0.6 & 0.5-0.4 & 0.7-0.4 & 0.7-0.2 \\ \midrule
% Validation Accuracy
%  & 72.85 & 73.21 & 73.66 & 72.35 \\
%  \bottomrule[0.4mm]
%     \end{tabular}
% 	\end{center}
%  \vspace{-1.5em}
% 	\label{table:abl}
% \end{table}

\subsection{Robustness against adversarial attacks} \label{sec:robustness}

As a by-product,  we would expect that DNNs trained within the CMIC-DL framework are more robust against adversarial attacks, in comparison with their counterparts trained within the standard CE-based DL framework. This is because when a DNN is trained within our CMIC-DL framework, its clusters in its output probability distribution space  are more compact, and also further separated from each other, in comparison with its counterpart trained within the standard CE-based DL framework. As such, it is harder for an adversary to craft a perturbation which,  when added to a clean sample, would  result in an  attacked sample falling into a cluster with a different label. Our purpose in this subsection is to confirm this by-product. To this end,  we have performed the following experiments.

 $\bullet$ \textbf{Dataset}: We have used MNIST dataset \cite{lecun1998gradient} comprising of 10-class handwritten digits. 

 $\bullet$ \textbf{Model}: We have selected a simple DNN with three convolutional and one fully connected layers.

 $\bullet$ \textbf{Attacks}:  Two white-box attacks have been selected,  where the adversary has an access to the gradients of the underlying model. Specifically, FGSM \cite{goodfellow2014explaining}  and PGD attack \cite{madry2018towards} with 5 iterations were employed with  attack perturbation budgets  $\| \epsilon \|_{\infty}=\{0.05,0.10,0.15,0.20,0.25,0.30,0.35\}$.

 $\bullet$ \textbf{Training settings}: 
We have deployed an SGD optimizer with a batch size of 64. We have trained the models for 15 epochs and adopted an step learning rate annealing with decay factor of 0.7. The hyper parameters were selected to be $\lambda^*=2$ and $\beta^* = 9$ in our CMIC-DL framework due to the fact that the classification task over MNIST dataset is far simpler than that over CIFAR-100 and ImageNet dataset.   

\begin{figure}[t] 
	\centering
	\subfloat[]{\includegraphics[width=0.5\columnwidth]{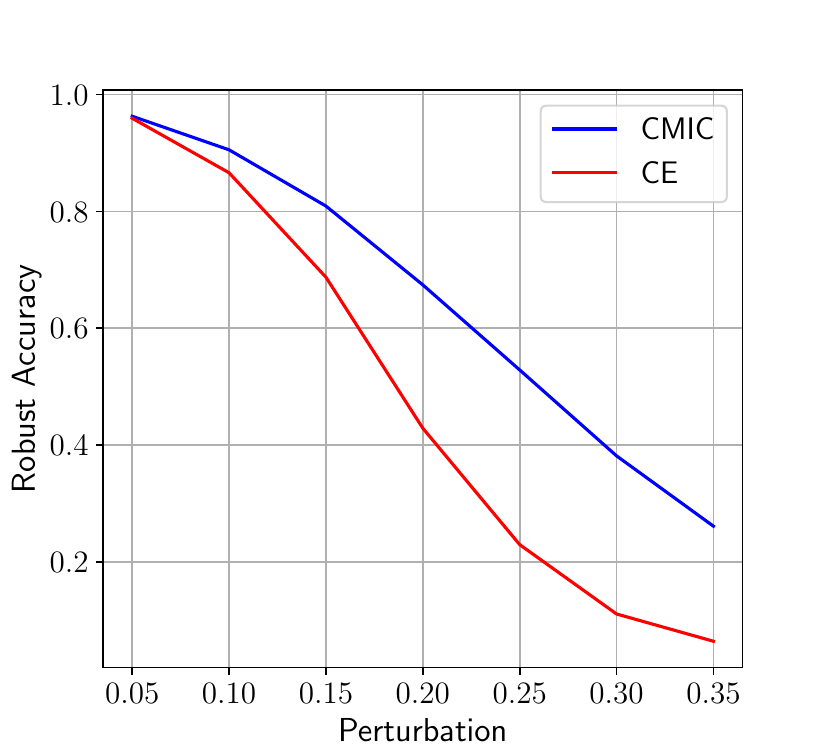}\label{L1}}
	%\vspace{-.15cm}
	\subfloat[]{\includegraphics[width=0.5\columnwidth]{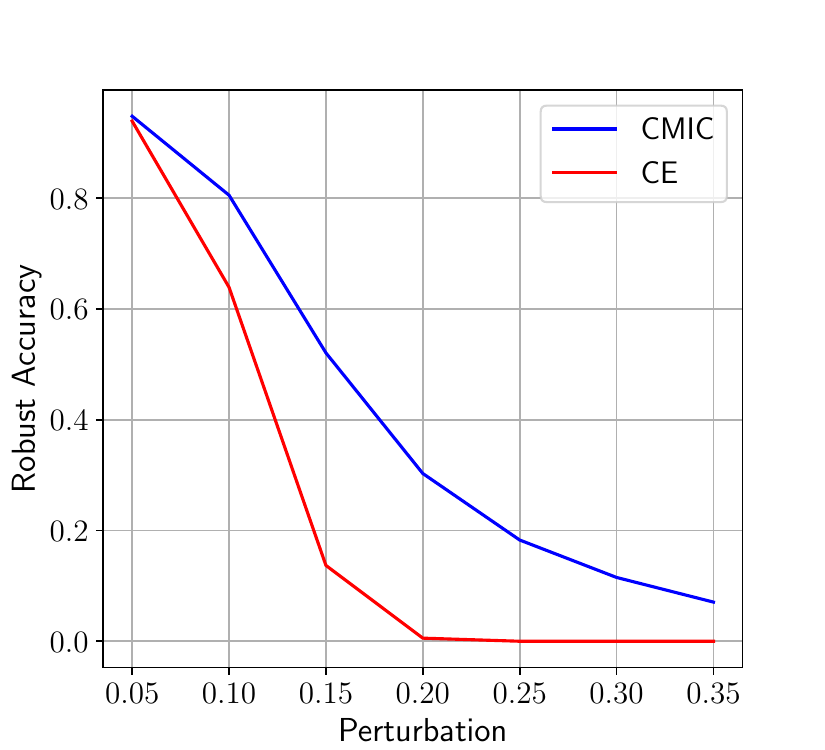}\label{L5}}
	\vspace{-.15cm}
    \caption{The robustness of a simple DNN over MNIST dataset trained within the conventional CE-based DL and CMIC-DL frameworks  against (a) FGSM attack and  (b) PGD attack with 5 iteraions, respectively.}
    \label{fig:attacks}
	\vspace{-0.6cm}
\end{figure}

Fig.~\ref{fig:attacks} illustrates the resulting trade-offs between robust accuracy and perturbation budget. From Fig. \ref{fig:attacks}, it is clear that the DNN trained within the CMIC-DL framework is more robust against both FGSM and PGD attacks, in comparison with its counterpart trained within the standard CE-based DL framework, thus confirming the by-product.  In addition, the clean accuracy for the models trained within the  CE-based DL and CMIC-DL frameworks are 99.14\% and 99.21\%, respectively, showcasing that the accuracy over the benign samples is not sacrificed for a higher robust accuracy.

We conclude this subsection by pointing out that although CMIC-DL can improve the robustness of DNNs trained therein against adversarial attacks, CMIC-DL itself is not a framework for adversarial training. In our future work, we will fully address CMIC adversarial training by extending the performance metrics of CMI, $\Gamma$ (separation), and NCMI to the new concepts of robust CMI, robust separation, and robust NCMI.

\section{Conclusion} \label{sec:con}

Viewing a DNN as a mapping from $x \in \mathbb{R}^{d} $  to $  P_x$, in this paper  we have introduced conditional mutual information (CMI) and normalized conditional mutual information (NCMI)  as new performance metrics of the DNN to measure the intra-class concentration and inter-class separation of the DNN. As new performance metrics, CMI and NCMI are in parallel with error rate. We then have used CMI and NCMI to evaluate and compare DNNs of different architectures and sizes. It turns out that NCMI and error rate have essentially a positive linear relationship with their correlation $\geq 0.99$. As such, the NCMI value of a DNN can be used to gauge the prediction performance of the DNN. 

Based on NCMI, we have then developed a learning framework called CMI constrained deep learning (CMI-DL) within which the conventional cross entropy function is minimized subject to a NCMI constraint. An novel alternating learning algorithm has been further proposed to solve such a constrained optimization problem. 
 Extensive experiment results consistently show that DNNs trained within the CMIC-DL framework outperform those trained using the other DL benchamrk methods discussed in the paper. In addition, with CMI and NCMI as performance metrics for measuring the concentration and separation of a DNN, the learning process of the DNN can also be analyzed and visualized through the evolution of CMI and NCMI.

 Open problems include (1) how to extend CMI and NCMI to define concepts of robust CMI, robust separation, and robust NCMI; (2) how to extend CMIC-DL to robust CMIC-DL to fully address adversarial training; (3) how to use CMI to help estimate the conditional probability distribution of $Y$ given $X$; and (4) the investigation of minimizing NCMI alone without using the standard cross entropy objective function by modifying a predictor. These problems will be addressed in the future.

\appendices
\setcounter{section}{0} 
\section{Proof of Theorem~\ref{th2}} \label{app-th2}

Since $\lambda >0 $ and $\beta >0$, it suffices to show that
 \begin{equation} \label{eq4-1}
     I(X; \hat{Y} | Y) = \min_{\{Q_c\}_{c \in [C]}} \be [  D (P_{X, \mathbf{\theta}} || Q_Y) ].
 \end{equation}
To this end, we apply \eqref{eq2-13} to get the following:
\begin{align}
& I(X; \hat{Y} | Y) = 
\sum_y \sum_{x } P(x, y )   \left [ \sum_{i=1}^C P(\hat{Y} = i |x,\mathbf{\theta} ) \times \right. \nonumber \\ 
& \left. \ln { P (\hat{Y} =i |x, \mathbf{\theta} ) \over P_{\hat{Y} | y } (\hat{Y} = i |Y =y )}  \right ] \nonumber \\ 
& = \sum_y \sum_{x } P(x, y )   \left [ \sum_{i=1}^C P(\hat{Y} = i |x, \mathbf{\theta}) \times  \left [ \ln { P (\hat{Y} =i |x, \mathbf{\theta} ) \over Q_y (i) } \right. \right. \nonumber \\
 & \left. \left.  +  \ln { Q_y (i)  \over P_{\hat{Y} | y } (\hat{Y} = i |Y =y )} \right ] \right ] \nonumber \\ 
 & = \sum_y \sum_{x } P(x, y )   D(P_{x,\mathbf{\theta}} || Q_y )   + \sum_y \sum_{x } P(x, y ) \times  \nonumber \\
 &    \left [ \sum_{i=1}^C P(\hat{Y} = i |x,\mathbf{\theta} )  \ln { Q_y (i)  \over P_{\hat{Y} | y } (\hat{Y} = i |Y =y )} \right ] \nonumber \\ 
 & = \be [  D(P_{X,\mathbf{\theta}} || Q_Y )]   \nonumber \\
 &  + \sum_y P(y )   \left [ \sum_{i=1}^C P_{\hat{Y} | y} (\hat{Y} = i |y  )  \ln { Q_y (i)  \over P_{\hat{Y} | y } (\hat{Y} = i |Y =y )}  \right ] \nonumber \\ 
 & = \be [  D(P_{X,\mathbf{\theta}} || Q_Y ) ] - \be [ D (P_{\hat{Y} | Y} || Q_Y) ] \nonumber \\
 & \leq  \be [  D(P_{X,\mathbf{\theta}} || Q_Y ) ],
\label{eq:CMI_2}  
\end{align}
for any  $Q_y \in {\cal P} ([C]), y \in [C]$, where the inequality above is due to the  nonnegativity of KL divergence. Thus
\begin{equation} \label{eq4-2}
     I(X; \hat{Y} | Y) \leq \min_{\{Q_c\}_{c \in [C]}} \be [  D (P_{X, \mathbf{\theta}} || Q_Y) ].
 \end{equation}
On the other hand, \eqref{eq:CMI_2} becomes an equality whenever 
\begin{equation}
   Q_c= P_{\hat{Y} | y=c }, \forall c \in [C]. \label{eq4-3}
\end{equation}
This, together with \eqref{eq:CMI_2}, implies \eqref{eq4-1}, and hence completes the proof of Theorem~\ref{th2}. 

\setcounter{section}{1} 
\section{Other Information Quantities for Separation} \label{app-sep}

In this Appendix, we explore other information quantities which can also be defined and used as a measure for the inter-class separation of the
DNN: $x \in \mathbb{R}^{d} \to  P_x$. Specifically,  two more information quantities $\Gamma^\prime$ and  $\Gamma''$ are introduced and compared with $\Gamma$ defined in  \eqref{eq2-17}. Although they are more or less equivalent, $\Gamma$ is more convenient for selecting hyper parameters in our CMIC-DL framework.

\subsection{Information Quantity $\Gamma^\prime$} \label{sec:gamma1}
A possible information quantity for measuring inter-class separation can be defined as follows
\begin{align} \label{eq:sep2}
\Gamma^\prime= \be \left [ I_{\{Y \not = V\}} D(P_{X} || P_{U}) \right ],
\end{align}
where the cross entropy function $H(P_{X}, P_{U})$ in \eqref{eq2-17} is replaced by the KL divergence $D(P_{X} || P_{U})$. To connect $\Gamma'$ with CMI and $\Gamma$, we simplify $\Gamma^\prime$ as follows:
\begin{align}  
\Gamma^\prime & =  \be \left[ I_{\{Y \not = V\}} \sum_{i=1}^C P(\hat{Y} = i |X)  \ln { \frac{ P (\hat{Y} =i |X )}{P (\hat{Y} =i |U )}} \right] \nonumber \\
& =  \be \left[ I_{\{Y \not = V\}} \sum_{i=1}^C P(\hat{Y} = i |X)  \left( \ln{\frac{P (\hat{Y} =i |X )}{P_{\hat{Y} | Y} (\hat{Y} = i |Y  )}} \right. \right. \nonumber \\
& \left. \left.  + \ln{\frac{P_{\hat{Y} | Y} (\hat{Y} = i |Y  )}{P (\hat{Y} =i |U )}} \right) \right] \nonumber \\
&= \be \left[  I_{\{Y \not = V\}}   D (P_X || P_{\hat{Y} |Y})  \right] \nonumber \\
& + \be \left[ I_{\{Y \not = V\}} \sum_{i=1}^C P(\hat{Y} = i |X) \ln{\frac{P_{\hat{Y} | Y } (\hat{Y} = i |Y  )}{P (\hat{Y} =i |U )}} \right] 
\label{eq:gam3}\\ 
& =  \be \left [ (1 - P(Y))  D (P_X || P_{\hat{Y} |Y})  \right ]  \label{eq:gam1} \\
& + \be  \left[ I_{\{Y \not = V\}} \sum_{i=1}^C P_{\hat{Y} | Y } (\hat{Y} = i |Y  ) \ln{\frac{P_{\hat{Y} | Y} (\hat{Y} = i |Y  )}{P (\hat{Y} =i |U )}} \right]  \label{eq:gam4}\\
& =  \be \left [ (1 - P(Y))  D (P_X || P_{\hat{Y} |Y})  \right ]  \nonumber \\
& + \be \left[  I_{\{Y \not = V\}} D(P_{\hat{Y} |Y}||P_U)\right], \label{eq:gam7}
\end{align}
where \eqref{eq:gam1} is due to the fact that $V$ is independent of $(X, Y)$, and \eqref{eq:gam4} follows from the independence of $(X, Y)$ and $(U, V)$ and the Markov chain $Y \to X \to \hat{Y}$.

Note that the first expectation in \eqref{eq:gam7} is related to the CMI $I(X; \hat{Y} |Y)$. Indeed, when $P(Y) $ is equal to a constant, i.e., $1/C$, which is true in most empirical cases,  it follows from \eqref{eq2-13} that 
 \[ \be \left [ (1 - P(Y))  D (P_X || P_{\hat{Y} |Y})  \right ] = (1 - {1 \over C}) I(X, \hat{Y} |Y) , \]
 which, together with \eqref{eq:gam7}, implies that
\begin{equation}
    \Gamma^\prime= (1 - {1 \over C}) I(X, \hat{Y} |Y) + \be \left[  I_{\{Y \not = V\}} D(P_{\hat{Y} |Y}||P_U)\right]. \label{eq:gam8}
\end{equation}
 Plugging \eqref{eq:gam8} into the optimization problem in \eqref{eq:uncons1}, we get the following optimization problem
\begin{align} \label{eq:unconsapp}
\min_{\mathbf{\theta} }~ \be_{X} & \left [ H (P_{Y|X},  P_{X,  \mathbf{\theta}}) \right ]+ \left( \lambda-\left(\beta-\frac{\beta}{C}\right)  \right) I(X; \hat{Y} | Y) \nonumber \\
&-\beta \be  \left [ I_{\{Y \not = V\}} D(P_{\hat{Y} |Y}||P_{U, \mathbf{\theta}}) \right ].
\end{align}
Thus, if $\Gamma^\prime$ was used as a measure for inter-class separation, then it would cancel out part of the CMI, making the selection of hyper parameters $\lambda$ and $\beta$ become harder.

\subsection{Information Quantity $\Gamma''$} \label{sec:gamma2}

Equations \eqref{eq:gam8} and \eqref{eq:unconsapp} suggest that one might use the following information quantity as a measure for inter-class separation instead 
\begin{align}\label{eq:gamma3}
\Gamma'' = \be \left[ I_{\{Y \not = V\}} D(P_{\hat{Y} |Y}||P_{U})\right] .
\end{align}
In fact, $\Gamma''$ has a descent physical meaning in the sense that it measures the average of distances between the output distributions of the DNN in response to input sample instances and the centroids of the clusters with different ground truth labels. 

To connect $\Gamma''$ with CMI and $\Gamma$, we further simplify $\Gamma''$ as follows
\begin{align}
\Gamma'' & = \be \left[ I_{\{Y \not = V\}} \sum_{i=1}^C P(\hat{Y} = i |X) \ln{\frac{P_{\hat{Y} | Y } (\hat{Y} = i |Y  )}{P (\hat{Y} =i |U )}} \right] 
\label{eq:gamma4}\\
& =  \be \left[ I_{\{Y \not = V\}} H( P_X, P_U) \right ] \nonumber \\
& + \be \left[ I_{\{Y \not = V\}} \sum_{i=1}^C P(\hat{Y} = i |X) \ln P_{\hat{Y} | Y } (\hat{Y} = i |Y  ) \right] \nonumber \\
& =  \Gamma  + \be \left[ I_{\{Y \not = V\}} \sum_{i=1}^C P_{\hat{Y}|Y} (\hat{Y} = i |Y) \ln P_{\hat{Y} | Y } (\hat{Y} = i |Y  ) \right] \label{eq:gamma5} \\
& = \Gamma - \be \left[ (1- P(Y)) H(P_{\hat{Y} |Y}, P_{\hat{Y} |Y}) \right ]. \label{eq:gamma6}  
\end{align}
In the above, \eqref{eq:gamma4} follows from \eqref{eq:gam3} and \eqref{eq:gam7}, \eqref{eq:gamma5} is due to the fact that $X$ is independent of $V$, and $Y \to X \to \hat{Y}$ forms a Markov chain, and \eqref{eq:gamma6} is attributable to the independence of $V$ and $Y$.

Note again that the second term  in \eqref{eq:gamma6} is related to the CMI $I(X; \hat{Y} |Y)$. Indeed, when $P(Y) $ is equal to a constant, i.e., $1/C$, which is true in most empirical cases,  it follows that 
\begin{eqnarray} \label{eq:gamma7}
\lefteqn{ \be \left[ (1- P(Y)) H(P_{\hat{Y} |Y}, P_{\hat{Y} |Y}) \right ] } \nonumber \\ 
& = &
    (1 - {1\over C}) H (\hat{Y} |Y ) \nonumber \\
& = &(1 - {1\over C})  \left [ I(X; \hat{Y} |Y) + H(\hat{Y} | X, Y) \right ] \nonumber \\
& = &(1 - {1\over C})  \left [ I(X; \hat{Y} |Y) + H(\hat{Y} | X ) \right ],
\end{eqnarray}
where $H(W |Z) $ denotes the Shannon conditional entropy of the random variable $ W$ given the random variable $Z$, and \eqref{eq:gamma7} is due to the Markov chain $Y \to X \to \hat{Y}$. Combining \eqref{eq:gamma7} with \eqref{eq:gamma6} yields 
\begin{equation} \label{eq:gamma8}
  \Gamma'' = \Gamma -   (1 - {1\over C})  \left [ I(X; \hat{Y} |Y) + H(\hat{Y} | X ) \right ].
\end{equation}

 Plugging \eqref{eq:gamma8} into the optimization problem in \eqref{eq:uncons1}, we get the following optimization problem
\begin{align} \label{eq:gamma9}
\min_{\mathbf{\theta} }~ \be_{X} & \left [ H (P_{Y|X},  P_{X,  \mathbf{\theta}}) \right ]+ \left( \lambda+\left(\beta-\frac{\beta}{C}\right)  \right) I(X; \hat{Y} | Y) \nonumber \\
& + \beta (1 - {1\over C})  H(\hat{Y} | X ) -\beta \Gamma . 
\end{align}
Thus, if $\Gamma''$ was used as a measure for inter-class separation, then it would further enhance the effect of the CMI, making the selection of hyper parameters $\lambda$ and $\beta$ become harder as well.

\section*{Acknowledgments}
This work was supported in part by the Natural Sciences and Engineering Research Council of Canada under Grant RGPIN203035-22, and in part by the Canada Research Chairs Program.

\bibliographystyle{IEEEtran}
\bibliography{egbib}

% \vspace{-1cm}
\begin{IEEEbiography}
[{\includegraphics[width=1in,height=1.25in,clip,keepaspectratio]{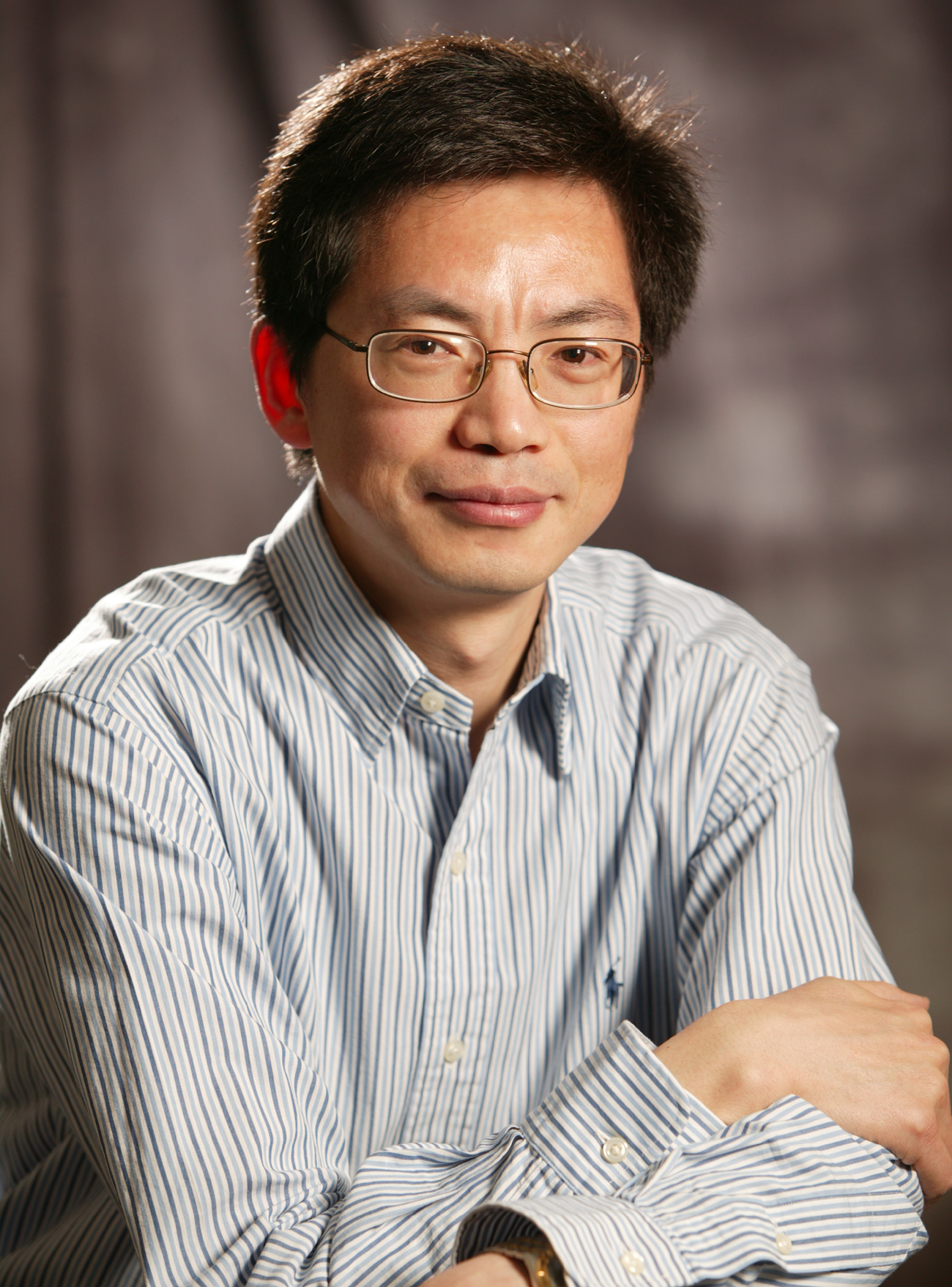}}]{En-Hui Yang} (M'97-SM'00-F'08) received the B.S. degree in applied mathematics from Huaqiao University, China, Ph.D. degree in mathematics from Nankai University, China, and Ph.D. degree in electrical engineering from the University of Southern California, USA, in 1986, 1991, and 1996, respectively.

Since June 1997, he has been with the Department of Electrical and Computer Engineering, University of Waterloo, ON, Canada, where he is currently a Professor and Canada Research Chair, and the founding Director of the Leitch-University of Waterloo multimedia communications lab. A co-founder of SlipStream Data Inc. (now a subsidiary of BlackBerry) and the founder of BicDroid Inc., he currently also serves as an Executive Council Member of China Overseas Friendship Association, an Expert Advisor for the Overseas Chinese Affairs Office of the State Council of China, a Board Governor of the University of Waterloo, a Board Trustee of Huaqiao University, a member of IEEE Founders Medal Committee, and advisors for other national and provincial bodies. His current research interests are: multimedia compression, digital communications, information theory, source and channel coding, image and video coding, deep learning, big data analytics, and information security.

Dr. Yang is a recipient of several awards and honors, a partial list of which includes the 2021 IEEE Eric E. Sumner Award, the prestigious Inaugural Premier's Catalyst Award in 2007 for the Innovator of the Year; the 2007 Ernest C. Manning Award of Distinction, one of the Canada's most prestigious innovation prizes; the 2013 CPAC Professional Achievement Award; the 2014 IEEE Information Theory Society Padovani Lecture; and the 2014 FCCP Education Foundation Award of Merit. With over 230 papers and more than 230 patents/patent applications worldwide, his research work has benefited people over 170 countries through commercialized products, video coding open sources, and video coding standards. He is a Fellow of the Canadian Academy of Engineering and a Fellow of the Royal Society of Canada: the Academies of Arts, Humanities and Sciences of Canada. He served, inter alia, as a review panel member for the International Council for Science; a general co-chair of the 2008 IEEE International Symposium on Information Theory; an Associate Editor for IEEE Transactions on Information Theory; a Technical Program Vice-Chair of the 2006 IEEE International Conference on Multimedia and Expo (ICME); the Chair of the award committee for the 2004 Canadian Award in Telecommunications; a Co-Editor of the 2004 Special Issue of the IEEE Transactions on Information Theory; and a Co-Chair of the 2003 Canadian Workshop on Information Theory.
\end{IEEEbiography}
\vspace{-1cm}

\begin{IEEEbiography}
[{\includegraphics[width=0.92in,height=1.25in,clip,keepaspectratio]{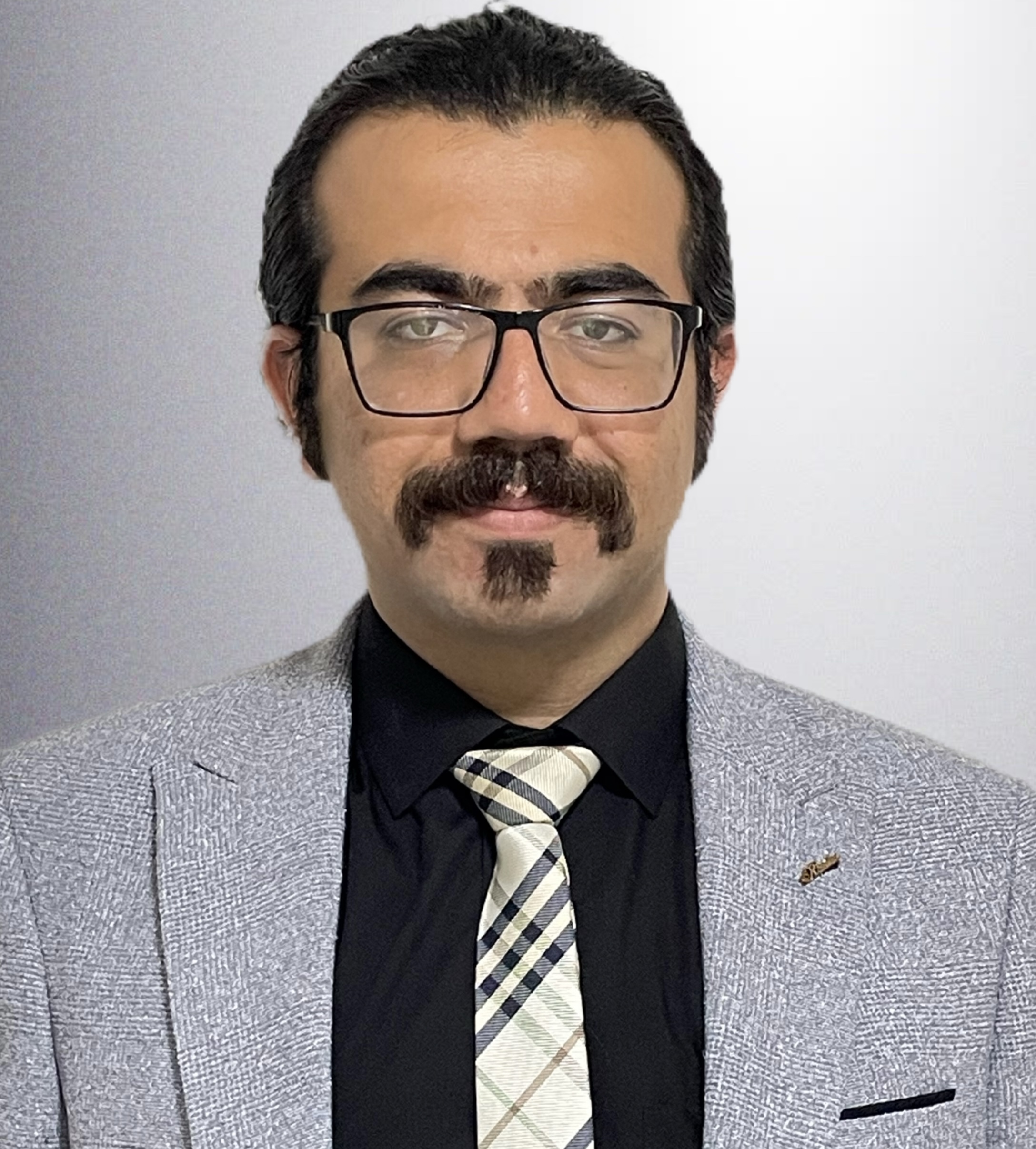}}]{Shayan Mohajer Hamidi} received the B.Sc. degree from Sharif University, Tehran, Iran, in 2016, and the MASc. degree from the University of Waterloo, Waterloo, ON, Canada, in 2018, both in electrical engineering. He was a research assistant with the CST lab at the University of Waterloo from 2018 to 2020. He is currently working toward his Ph.D. degree in Electrical Engineering at the University of Waterloo. His current research interests include machine learning, optimization, information theory.
\end{IEEEbiography}

\vspace{-1cm}
\begin{IEEEbiography}
[{\includegraphics[width=0.92in,height=1.25in,clip,keepaspectratio]{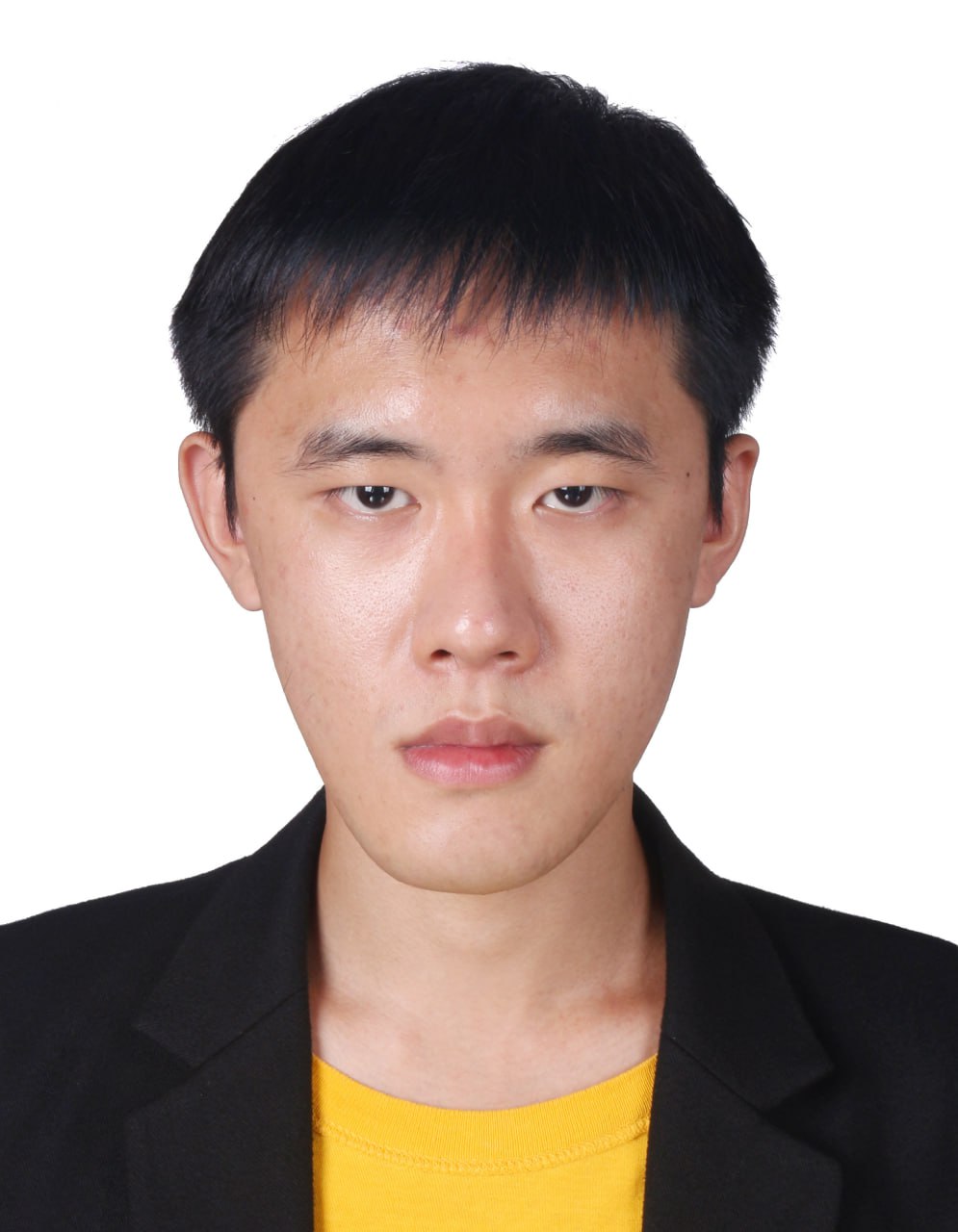}}]{Linfeng Ye} received the B.Sc. degree from Xi'an University of Science and Technology, Xi'an, China in 2020, in microelectronics, and the MEng. degree from the University of Waterloo, Waterloo, ON, Canada, in 2021, in electrical engineering. He is currently working towards his MASc. degree in electrical engineering at University of Waterloo. His current research interests include machine learning, and information theory.
\end{IEEEbiography}

\vspace{-1cm}
\begin{IEEEbiography}
[{\includegraphics[width=1in,height=1.25in,clip,keepaspectratio]{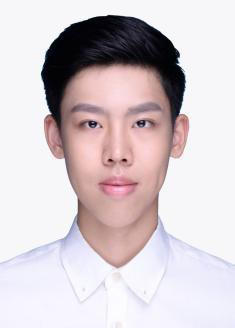}}]{Renhao Tan} received the Bachelor of Advanced Computing (Honours) degree from the Australian National University, Canberra, ACT, Australia, in 2021. He is currently pursuing a MASc. degree in Electrical \& Computer Engineering at University of Waterloo, Waterloo, ON, Canada. His research interests include machine learning, computer vision and data mining.
\end{IEEEbiography}

\begin{IEEEbiography}
{Beverly Yang} received the BASc degree in Geological Engineering from the University of Waterloo in 2020. She is currently pursuing a PhD in mining engineering at the University of British Columbia. Her research interests include rock mechanics and machine learning.
\end{IEEEbiography}

\end{document}